\documentclass[conference]{IEEEtran}
\usepackage{times}
\usepackage{amsmath,amssymb,amsfonts}
\usepackage{algorithmic}
\usepackage{graphicx}
\usepackage{textcomp}
\usepackage{xcolor}
\usepackage[numbers]{natbib}
\usepackage{multicol}
\usepackage[bookmarks=true]{hyperref}

\begin{document}

\title{Tactile sensing enables vertical obstacle negotiation for elongate many-legged robots}



%
\author{\authorblockN{Juntao He\authorrefmark{1},
Baxi Chong\authorrefmark{2},
Massimiliano Iaschi\authorrefmark{3},
Vincent R. Nienhusser\authorrefmark{3}, 
Sehoon Ha\authorrefmark{4} and Daniel I. Goldman \authorrefmark{2}}
\authorblockA{\authorrefmark{1}Institute for Robotics and Intelligent Machines\\ Georgia Institute of Technology, Email: jhe391@gatech.edu}
\authorblockA{\authorrefmark{2}School of Physics, Georgia Institute of Technology}
\authorblockA{\authorrefmark{3}School of Mechanical Engineering, Georgia Institute of Technology }
\authorblockA{\authorrefmark{4} College of Computing, Georgia Institute of Technology}}


\maketitle

\begin{abstract}
Many-legged elongated robots show promise for reliable mobility on rugged landscapes. However, most studies on these systems focus on planar motion planning without addressing rapid vertical motion. Despite their success on mild rugged terrains, recent field tests reveal a critical need for 3D behaviors (e.g., climbing or traversing tall obstacles). The challenges of 3D motion planning partially lie in designing sensing and control for a complex high-degree-of-freedom system, typically with over 25 degrees of freedom. To address the first challenge regarding sensing, we propose a tactile antenna system that enables the robot to probe obstacles to gather information about their structure. Building on this sensory input, we develop a control framework that integrates data from the antenna and foot contact sensors to dynamically adjust the robot's vertical body undulation for effective climbing. With the addition of simple, low-bandwidth tactile sensors, a robot with high static stability and redundancy exhibits predictable climbing performance in complex environments using a simple feedback controller. Laboratory and outdoor experiments demonstrate the robot's ability to climb obstacles up to five times its height. Moreover, the robot exhibits robust climbing capabilities on obstacles covered with shifting, robot-sized random items and those characterized by rapidly changing curvatures. These findings demonstrate an alternative solution to perceive the environment and facilitate effective response for legged robots, paving ways towards future highly capable, low-profile many-legged robots.

\noindent Page: \textcolor{magenta}{\url{https://juntaohe.github.io/Climbing_RSS2025/}
}
\end{abstract}

\IEEEpeerreviewmaketitle

\section{Introduction}
\begin{figure}[ht]
    \centering
    \includegraphics[width=8.5cm]{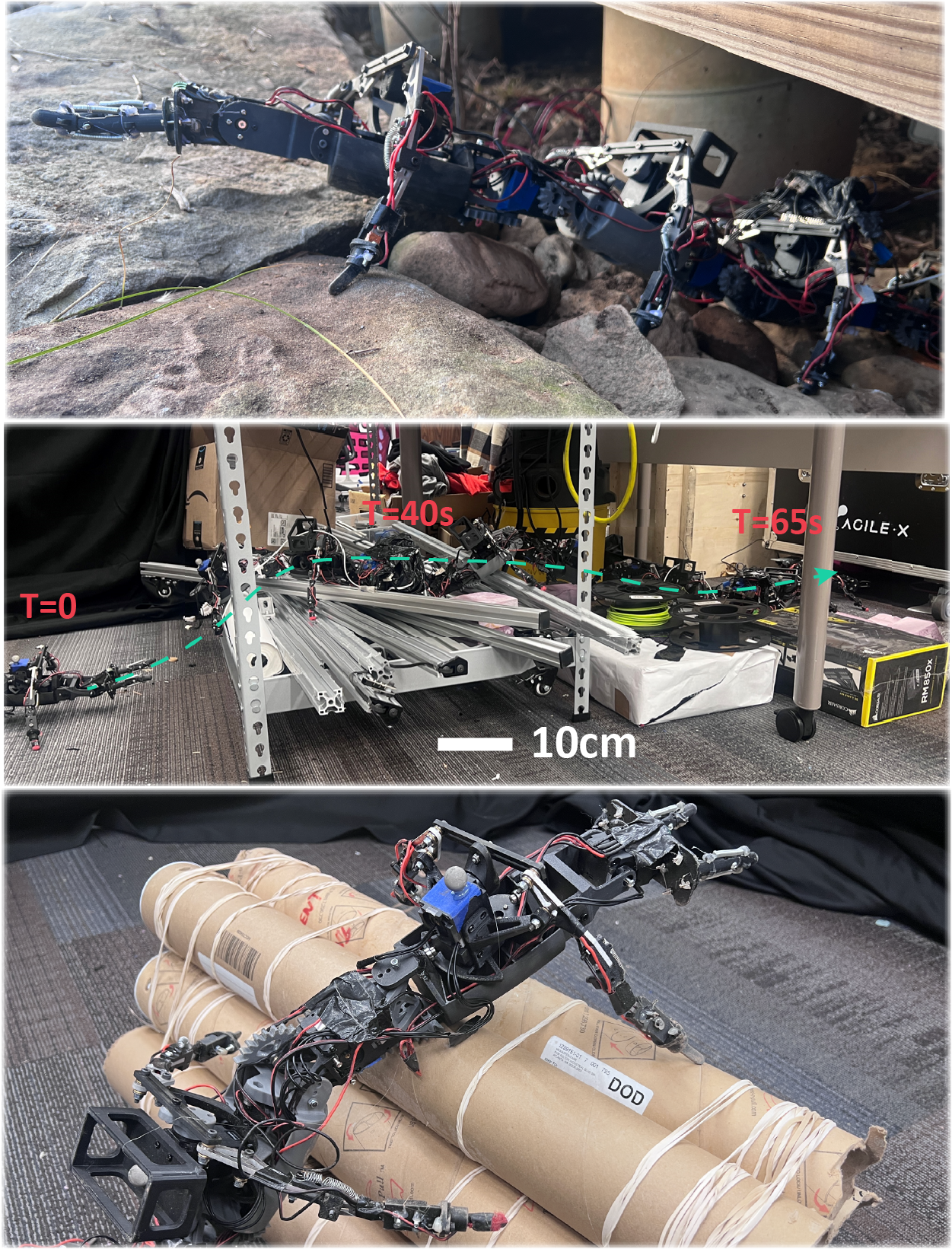}
    \caption{\textbf{Tacitly sensing multilegged robot climbing in different environments}. (Top) The robot climbs a large rock, four times its height, in a confined space with terrain covered in mud, grass, and scattered boulders. (Middle) The robot successfully navigates confined environments with vertical obstacles, unstable metal bars, and flowable plastic disks. (Bottom) The robot climbs obstacles five times its height with rapidly changing curvatures in a laboratory setting.}
    \label{fig:1}
\end{figure}
Mid-sized elongated many-legged robots, ranging from 5 to 10 cm in height and 1 to 2 kg in weight, demonstrate exceptional mobility in navigating confined spaces and challenging terrains. For instance, studies \cite{he2024probabilistic,he2024learning} have shown that these robots can traverse environments featuring pine straw, robot-sized rocks, mud, bushes, and leaves. Despite these advancements, most research on these systems primarily focuses on 2D planar motion planning \cite{he2024probabilistic,he2024learning,teder2024effective,chong2023multilegged,chong2023self,chong2022general,aoi2022advanced,miyamoto2021analysis,yasui2017decentralized}. 

Achieving 3D behaviors, such as climbing or steep slope ascent, is essential, as it significantly broadens the range of environments these robots can navigate. This is particularly important for mid-sized robots navigating among real-world obstacles (e.g., rocks with height $10-30$cm, metal wreckage with height $20-40$cm, and boxes/tubes with height $20-30$cm, Fig.\ref{fig:1}). 

Despite the challenges in many-legged robots, extensive climbing studies have been conducted on other few-legged robots\cite{hoeller2024anymal,luo2024pie,cheng2024extreme,haynes2009rapid,nagakubo1994walking,saunders2006rise,spenko2008biologically}. Quadrupedal robots accomplish impressive parkour tasks by incorporating vision into their control frameworks \cite{hoeller2024anymal,luo2024pie,cheng2024extreme}. Hexapod and quadrupedal robots utilizing adhesion mechanisms, such as vacuum pumps \cite{yan1999development,gao2004study,miyake2009vacuum}, magnetic adhesion \cite{gao2008boiler,lee2012combot,hong2022agile}, and bio-inspired claws \cite{saunders2006rise,spenko2008biologically,haynes2009rapid,bretl2006motion,parness2017lemur,chen2024locomotion}, have demonstrated robust climbing capabilities on vertical walls and trees. 

However, directly transferring the few-legged climbing behaviors to many-legged robots presents substantial challenge.
The first challenge is the vision quality. Climbing in quadrupedal robots relies heavily on camera data to reconstruct the geometric features of obstacles for motion planning \cite{qi2021perceptive,hoeller2024anymal,cheng2024extreme}. The vision in many-legged robots suffers from the inherent poor quality because (i) many-legged robots often operate in confined environments where low-light conditions obstruct vision \cite{ying2017new,rashed2019fusemodnet,zamir2021learning}, (ii) many-legged robots experience substantial body orientation oscillation during locomotion, which significantly compromise the focus of the vision, and (iii) the position of camera is typically positioned too close to the ground, limiting the ability to gather whole-terrain information. The second challenge is the adhesion design. Adhesion-based robots are constrained by their specialized foot designs, which limit their deployment to specific surfaces, such as metal pipes \cite{gao2008boiler,lee2012combot,hong2022agile} or smooth glass walls \cite{yan1999development,gao2004study,miyake2009vacuum}. To enable many-legged robots to traverse more general terrains, adhesive foot designs are impractical for enhancing climbing capabilities. Finally, 3D motion planning is particularly challenging for these complicated high degree-of-freedom (DoF) systems, which have over 25 DoF. That is, the complexity of their dynamic, whole-body interactions with the environment poses additional difficulties in developing robust control strategies. Partially because of a lack of systematic locomotion research on their biological counterparts, we have limited intuition on ``what to sense" and ``how to respond" in many-legged systems.

In addition to vision, tactile sensing offers a reliable short-range perception framework \cite{charlebois1996curvature,fearing1988using}. Recent studies \cite{wu2019tactile,li2024whisker,berman2024additively,pezzementi2011object} highlight its ability to accurately estimate the geometric features of objects at close range. Compared to vision-based approaches, tactile sensing provides advantages such as lower computational requirements and insensitivity to lighting conditions, despite its limited detection range. 

Building on these advantages, recent studies have integrated tactile sensing into mid-sized robotic systems for tasks such as environmental geometry estimation and terrain roughness assessment. Bio-inspired tactile antennas \cite{mongeau2014mechanical,mongeau2015sensory,lee2008templates,lewinger2005insect} have been successfully incorporated into robotic control frameworks, enabling behaviors like wall-following and climbing. Snake robots have demonstrated impressive mobility on complex terrain using tactile sensing \cite{fu2023contact,ramesh2022sensnake}. Similarly, tactile sensing has been integrated into hexapod systems \cite{mrva2015tactile,luneckas2021hybrid,wu2019tactile} to support effective gait adaptation in challenging environments. Many-legged robots \cite{he2024probabilistic,he2024learning,he2024tactile} have also employed tactile foot contact sensors to adapt their gaits, showing significant improvements in speed over rugged terrain. These advances suggest that a tactile sensory system is particularly well-suited for many-legged robots performing climbing tasks, especially in highly rugged and low-light environments.

In this work, we demonstrate that a mechanically intelligent many-legged robot, characterized by high static stability and redundancy, achieves predictable climbing performance in highly complex environments by integrating simple, low-bandwidth tactile sensors with a simple feedback controller.

First, we propose a tactile antenna system for short-range (10 cm) contact sensing to reconstruct obstacle geometry. We then investigate control algorithms for robot climbing, including open-loop and feedback-based approaches. Open-loop control, which coordinates limb stepping with horizontal and vertical body undulation, effectively handles obstacles up to twice the robot's height but has limited capability. To overcome these, we develop a feedback control framework that integrates antenna data and foot contact sensors. This controller raises the robot's head upon detecting an obstacle, pitches it downward to position it on top, and guides it along the obstacle’s contour using antenna contact data. Additionally, it accelerates the transition of floating segments to stable positions through pitch-down motions based on duty factor data from foot contact sensors.

Laboratory and outdoor experiments demonstrate the robot’s ability to climb obstacles up to five times its center height. Moreover, it successfully navigates obstacles with rapidly changing curvatures and those covered with shifting, robot-sized debris. We also validate the robustness of this climbing controller in complex outdoor environments: the robot successfully completed a pipe inspection within a 20 cm radius pipe filled with robot-sized rocks and leaves and climbed out of a 30 cm gap under a bridge, maneuvering over scattered giant rocks, dense weeds, and other vegetation. 

\section{Background: Wave templates in many-legged robot }
\label{wave pattern}
\begin{figure}[ht]
    \centering
    \includegraphics[width=9cm]{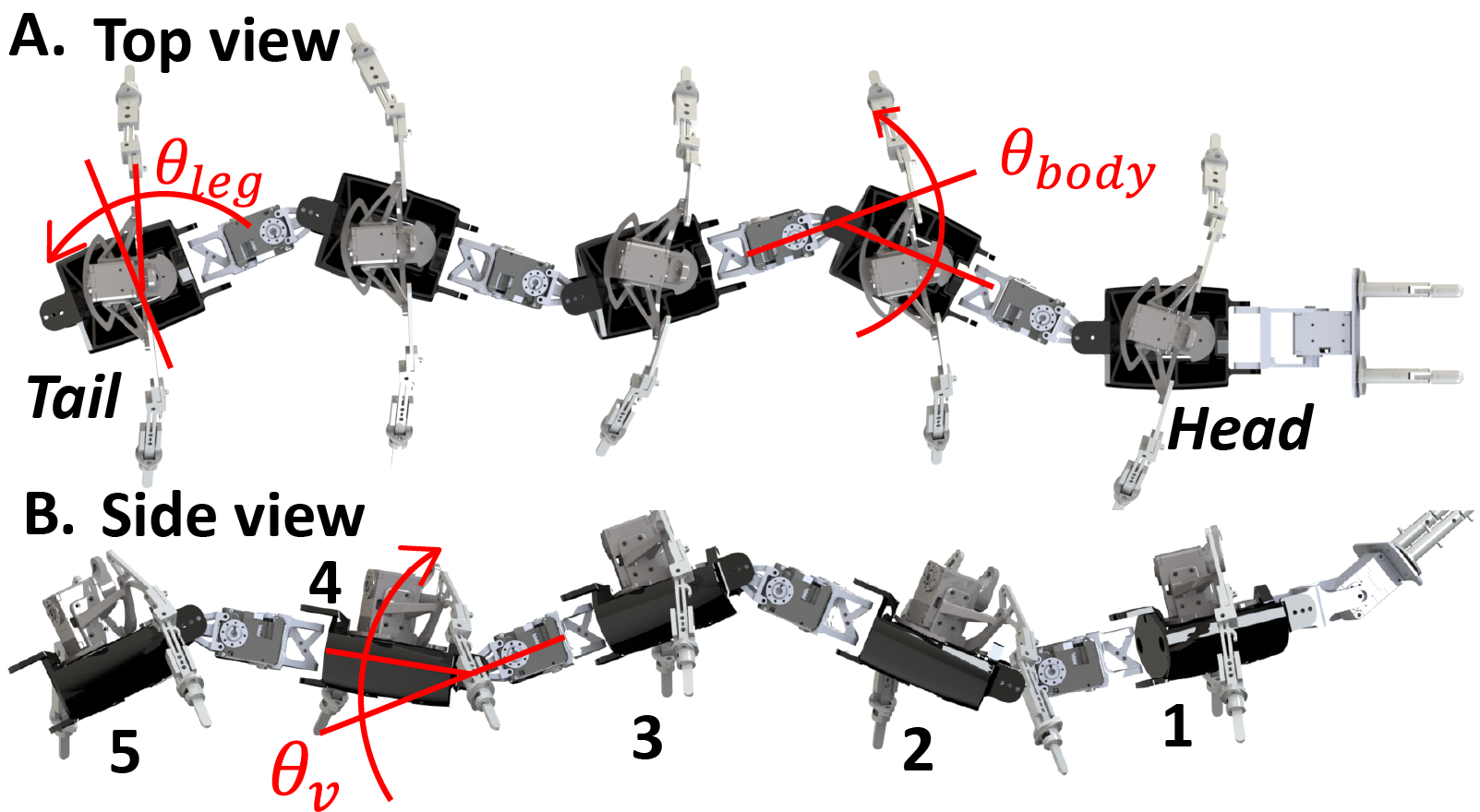}
    \caption{\textbf{Robot wave templates}: A. Overhead view of the robot: $\theta_{leg}$ (shoulder angle) and $\theta_{body}$ (horizontal body joint angle) are determined by leg amplitude $\Theta_{leg}$ and body amplitude $\Theta_{body}$, respectively. B. Side view of the robot: $\theta_v$ (vertical body joint angle) is determined by vertical amplitude $A_v$.}
    \label{fig:wave_pattern}
\end{figure}
 Previous studies \cite{chong2022general,chong2023multilegged,chong2023self,he2024probabilistic,he2024learning} have demonstrated that successful navigation of many-legged robots over rough terrain relies on coordinating leg movements with horizontal and vertical body undulations (Fig. \ref{fig:wave_pattern}). Specifically, forward motion is achieved by prescribing leg stepping and body undulation patterns as sinusoidal traveling waves. In this work, we utilize the wave template presented in this section to design an open-loop controller. For the feedback controller, we independently regulate the pitch motion of two vertical joints while the remaining joints follow the wave templates.

The robot's legs provide propulsion by retracting during the stance phase to maintain ground contact and protracting during the swing phase to disengage. During the stance phase, each leg moves from the anterior extreme to the posterior extreme, reversing direction during the swing phase. The anterior and posterior excursion angles ($\theta_{leg}$) for a given contact phase ($\tau_c$) are modeled using a piecewise sinusoidal function:
\begin{align}
        \theta_{leg,l}(\tau_c,1)  &=\begin{cases}
      \Theta_{leg}\cos{(\frac{\tau_c}{2D})}, & \text{if}\ \text{mod}(\tau_c,2\pi)  < 2\pi D\\
      -\Theta_{leg}\cos{(\frac{\tau_c-2\pi D}{2(1-D)})}, & \text{otherwise},
    \end{cases} \nonumber \\
    \theta_{leg,l}(\tau_c, i) &= \theta_l(\tau_c - 2\pi\frac{\xi}{n}(i-1), 1) \nonumber \\
    \theta_{leg,r}(\tau_c, i) &= \theta_l(\tau_c + \pi, i) 
    \label{eq:legmove}
\end{align}
where $\Theta_{leg}$ is the maximum shoulder angle, and $\theta_{leg,l}(\tau_c,i)$ and $\theta_{leg,r}(\tau_c,i)$ are the shoulder angles for the $i$-th left and right leg, respectively, at contact phase $\tau_c$. The shoulder angle peaks ($\theta_{leg}=\Theta_{leg}$) when transitioning from swing to stance and reaches its minimum ($\theta_{leg}=-\Theta_{leg}$) during the reverse transition. Unless otherwise specified, $D$ is assumed to be 0.5.

Horizontal body undulation is introduced by propagating a wave along the robot’s body from head to tail: \begin{align} \theta_{body}(\tau_b,i)=\Theta_{body} \cos(\tau_b - 2\pi\frac{\xi^b}{n}(i-1)), \label{eq
} \end{align}
\begin{figure}[h]
    \centering
    \includegraphics[width=7cm]{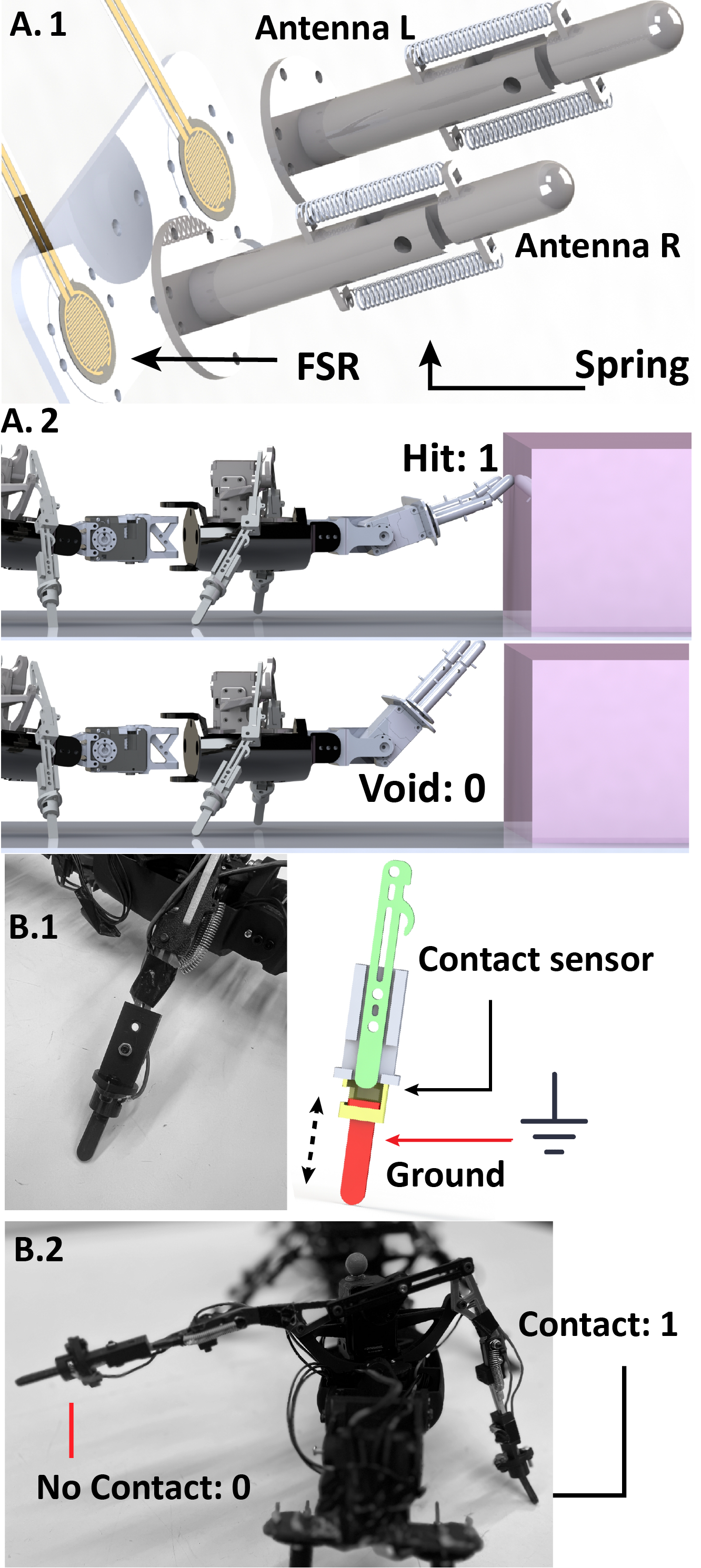}
    \caption{\textbf{Tactile sensory system.} \textbf{A. Antenna design}. A1. The base of the antenna is attached to a Force Resistive Sensor (FSR) using screws. The inner part of the antenna is connected to its tip with two springs, allowing it to deform upon contact with obstacles, thereby preventing jamming. A2. Antenna contact states: A value of 0 represents no contact (void state), while 1 indicates an obstacle has been detected. \textbf{B. Binary limb contact sensing system.}  B1. Design of a binary contact sensor for each foot, based on capacitive sensing. B2. Contact state of the leg: 0 indicates no contact, while 1 indicates contact.   
    }
    \label{fig:3}
\end{figure}
where $\theta_{body}(\tau_b,i)$ is the angle of the $i$-th body joint at phase $\tau_b$, and $\xi^b$ indicates the number of spatial waves along the body. For simplicity, we assume that the number of spatial waves in both body and leg movements is equal ($\xi^b = \xi$), allowing the lateral body wave to be parameterized by its phase $\tau_b$.

The gait of the many-legged robot is determined by the phases of contact ($\tau_c$) and lateral body undulation ($\tau_b$), combining leg and body waves. Effective coordination, ensuring proper leg retraction during motion, is achieved when $\tau_c = \tau_b - (\xi/N + 1/2)\pi$.

Vertical body undulation is introduced via a wave propagated along the robot’s backbone: \begin{align} \theta_v(\tau_b,i)=A_v \cos(2\tau_b - 4\pi\frac{\xi^b}{n}(i-1)), \label{eq
} \end{align} 
where $\theta_v(\tau_b,i)$ represents the vertical angle of the $i$-th body joint at phase $\tau_b$, and $A_v$ defines the wave’s amplitude.


\section{Tactile sensory systems}
In this section, we introduce two sensory mechanisms integrated into the climbing feedback control framework. The first mechanism, the antenna, estimates the obstacle's geometry using its hit data. The second mechanism, the foot contact sensor, detects the binary foot contact state and monitors the floating state of each robot segment.

\subsection{Tactile Antenna design}
\label{antenna_design}
We developed an antenna equipped with two Force Sensing Resistor (FSRs) sensors capable of detecting forces within a range of 0 to 10 N. As shown in Fig. \ref{fig:3}.A.1, each FSR is mounted on a flat surface, with the base of the antenna attached to the opposite side of the FSR using screws. The inner part of the antenna is connected to its tip via two springs, allowing it to deform upon contact with obstacles, thereby preventing jamming. According to our experiments, the addition of the antenna does not limit the robot’s physical capabilities, such as speed or maneuverability.

The contact states of the antenna are expressed using a binary system. When the antenna touches an obstacle, it is in the ``Hit'' state, represented by 1; otherwise, it is in the ``Void'' state, represented by 0. The FSR transmits an analog signal ranging from 0 to 1024 to the onboard controller, based on the magnitude of the force detected. In this work, we define a ``Hit" state (1) for analog values exceeding 300, corresponding to approximately 1 N, and a ``Void" state (0) for values below this threshold. 

\subsection{Binary limb contact sensing system design}
We implemented a low-bandwidth binary contact sensor system (Fig. \ref{fig:3}.B) to monitor foot-ground interaction for each leg, allowing us to compute duty factor of each robot segment and using this information as a feedback into our control system. Contact capacitive sensors (MPR121) embedded at the tip (highlighted in green in Fig.\ref{fig:3}.B.1) of each leg detect capacitance variance. The toe (highlighted in red in Fig.\ref{fig:3}.B.1) has a slight range of linear motion, resulting in minimal capacitance when the leg is suspended and maximal capacitance when it is grounded. The analog value shows a significant difference between the suspended state (greater than 200) and the grounded state (less than 5). Therefore, we classify any analog value below 50 as indicating contact.

\section{Control framework for climbing}
This section first examines the limitations of an open-loop controller, which coordinates limb stepping and horizontal/vertical body undulation wave patterns using the templates described in Section \ref{wave pattern}. The open-loop controller is limited to overcoming obstacles up to 10 cm in height, equivalent to twice the robot's height. To address this limitation, we propose a feedback controller that adjusts vertical joint motion by integrating antenna data and foot contact sensor inputs. This feedback controller extends the climbing capability to obstacles up to five times the robot's height. We further demonstrate the robustness of this controller in highly unstructured lab-based and outdoor environments.

\subsection{Open-loop control}
The robot's belly is naturally elevated above the ground, allowing it to overcome low obstacles (with obstacle height lower than the belly height, $h_{belly}=5$ cm) with the directionally compliant limbs~\cite{ozkan2020systematic}. 

\begin{figure}
    \centering
    \includegraphics[width=9cm]{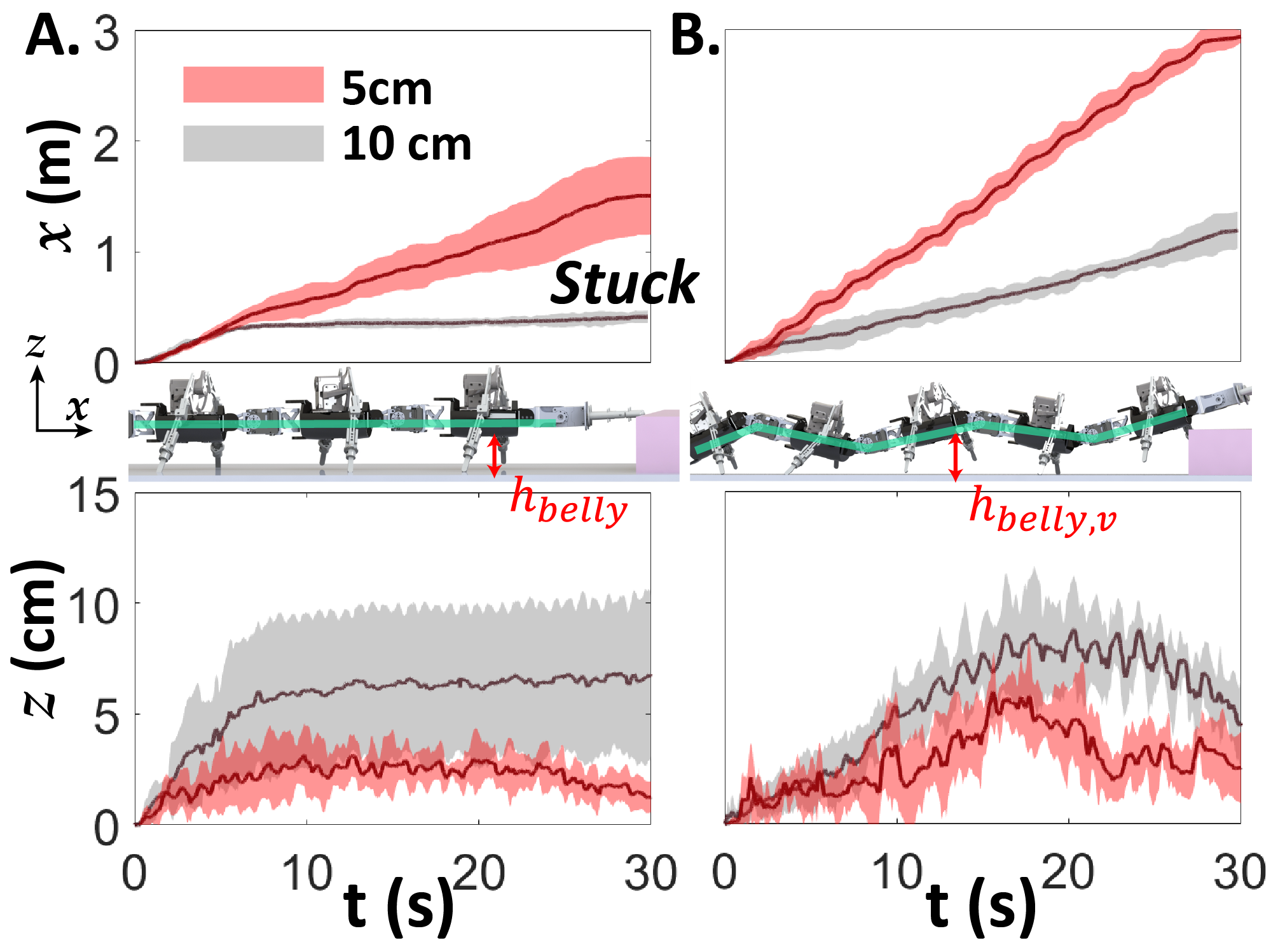}
    \caption{\textbf{Open loop experiments}. \textbf{A.} Forward/vertical displacement vs. time plot showing the robot climbing 5 cm and 10 cm obstacles without vertical body undulation. The inset illustrates how the robot becomes blocked without the use of vertical waves. \textbf{B.} A forward/vertical displacement vs. time plot showing the robot climbing 5 cm and 10 cm obstacles with vertical body undulation. The inset demonstrates how introducing vertical body undulation raises the robot's belly height, increasing its maximum climbing capability.}
    \label{fig:5}
\end{figure}

To test this, we evaluated the robot locomotion performance on box-shaped obstacles with heights of 5 cm and 10 cm. The obstacle dimensions are 45 cm in width and 120 cm in length, with the robot moving forward along the width direction. In the experiments, the robot was positioned 5 cm in front of the obstacle and allowed to run for 10 motion cycles (a total of 30 seconds). The motion was tracked using the Opti-Track motion tracking system. Climbing was defined as successful if the robot’s head fully passed to the other side of the obstacle. Experimental results showed that the robot successfully traversed the 5 cm obstacle. However, when attempting the 10 cm obstacle, the robot could only place its first two segments on the obstacle before becoming stuck (Fig.\ref{fig:5}.A). Here we set $\Theta_{body}=\pi/18$ and $\Theta_{leg}=\pi/6$. Our choice of gait parameter is based on the preliminary work~\cite{chong2023multilegged,chong2023self,he2024probabilistic} and empirical experiments. 

To overcome higher obstacles, and building on previous success in using vertical waves in rough terrain~\cite{chong2023multilegged,he2024probabilistic}, we propose utilizing vertical body undulation, which periodically raises a portion of the body. We tested our vertical wave modulation on robot experiments.
Introducing vertical body undulation into the gait raises the belly height ($h_{belly,v}$), thereby increasing the expected maximum climbing height. As discussed in Section \ref{wave pattern}, the maximum vertical joint angle is controlled by adjusting the vertical amplitude $A_v$, making $h_{belly,v}$ dependent on $A_v$. The maximum $A_v$ that ensures stable forward motion is $2\pi/9$, corresponding to a max($h_{belly,v}$) of 9 cm. 

We conducted experiments on obstacles with heights of 5 cm, 10 cm, and 15 cm using the gait with vertical body undulation. The results show that the robot successfully traverses obstacles up to 10 cm in height but is fully blocked by the 15 cm obstacle. 

In summary, the robot using open-loop control can traverse obstacles up to a height of 10 cm (Fig.\ref{fig:5}).
\subsection{Feedback control}
To increase the robot's maximum climbing height, we developed a control framework that independently controls the vertical motion of up to two body joints, instead of coupling the entire vertical body undulation to a single sinusoidal wave, as described in Section \ref{wave pattern}.

\begin{figure}
    \centering
    \includegraphics[width=8.5cm]{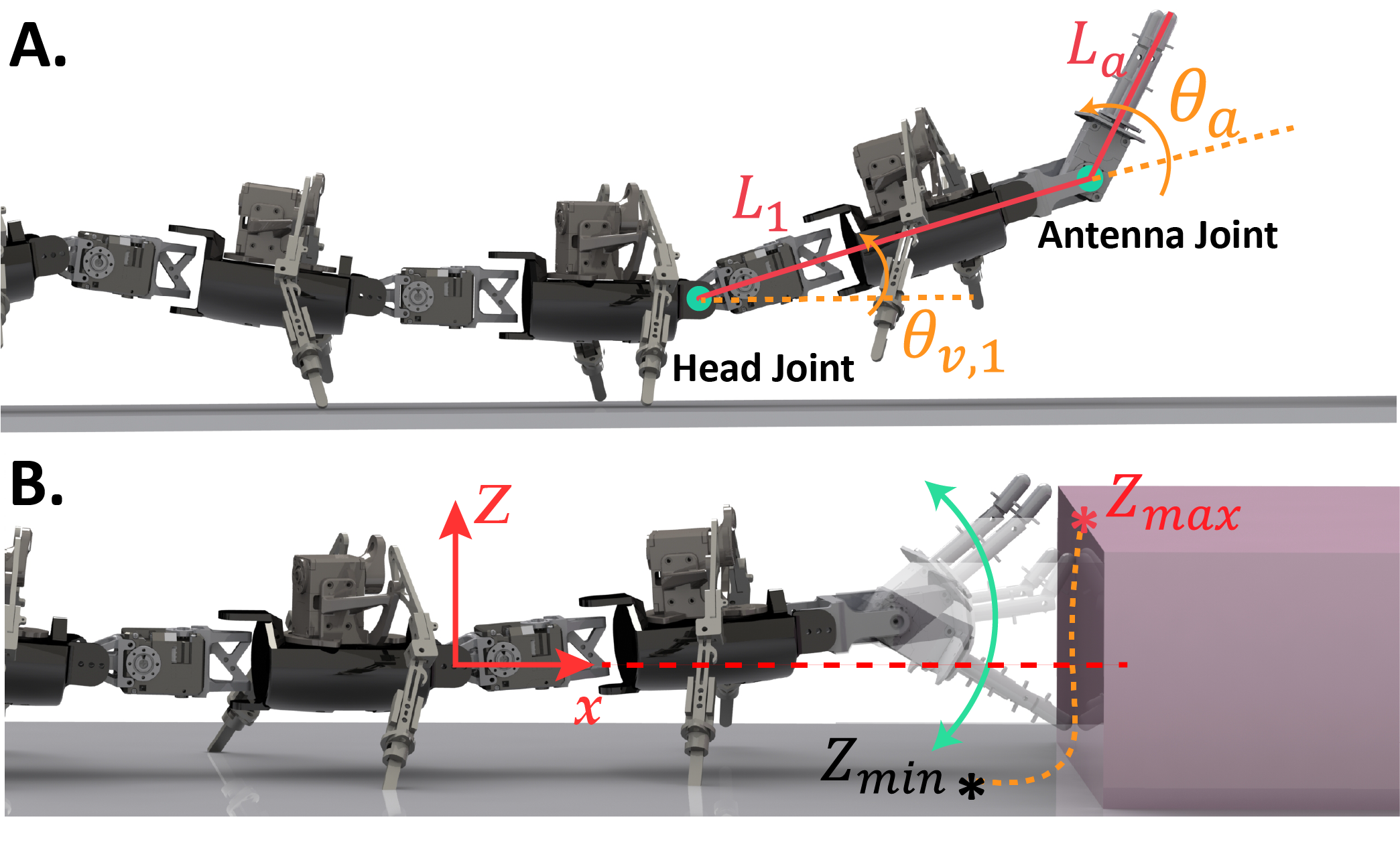}
    \caption{\textbf{Obstacle height estimation}. The z-position of the antenna tip relative to the head joint can be estimated using a rigid transformation, based on the length of the head segment ($L_1$) and the antenna ($L_a$), along with the joint angle history.}
    \label{fig:6}
\end{figure}
\subsubsection{Obstacle height estimation using antenna}
Climbing requires an estimation of an obstacle's geometric features. Here, we use the antenna introduced in Section \ref{antenna_design} to estimate obstacle height for subsequent robot control.

The antenna is programmed to oscillate vertically, with its joint angle following a sinusoidal function, \( A_a \sin(w_a t) \), where \( A_a \) represents the oscillation amplitude and \( w_a \) is the temporal frequency. In this study, \( A_a \) is set to \( 5\pi/18 \), ensuring the antenna can touch the ground when the robot operates on flat terrain. The temporal frequency \( w_a \) is set to 4, enabling the antenna to complete one full oscillation cycle within 1/4 of the robot's gait cycle. The robot's vertical motion is also adjusted at this frequency.

As shown in Figure \ref{fig:6}, we use the antenna's ``Hit" (1) and ``Void" (0) information (Fig. \ref{fig:3}.A.2) to estimate the obstacle's position relative to the head segment joint. An array $H = [h(t-T),h(t-T+1),...,h(t)]$ is created to store antenna's contact history, where $T$ represents the time interval for sampling the antenna's hit information. For example, at time $t$, if the antenna detects a hit, then $h(t)=1$ is appended to the contact history array $H$. Otherwise, $h(t)=0$ is appended to $H$. This process generates a binary sequence that records the antenna's hit history over time.

Using the robot's dimensions and the joint angle history of the head segment's vertical joint and the antenna joint, the Z-position of the antenna tip relative to the head joint is approximated via a rigid transformation (Fig. \ref{fig:6}):  
\begin{equation}
        z(t) = L_{1} \sin(\theta_{v,1}(t)) + L_{a} \sin(\theta_{v,1}(t)+\theta_a(t)),
\end{equation}
where \( L_1 \) and \( L_a \) are the lengths of the head segment and the antenna, respectively, and \( \theta_{v,1} \) and \( \theta_{a} \) are their corresponding joint angles. Since the length ratio between the tip and the base is less than 1:5, the error caused by tip deformation can be safely ignored. Similar to $H$, an array $Z = [z(t-T),z(t-T+1),...,z(t)]$ is created to store the Z-position history of the antenna. 

By combining the antenna contact history, \( H \), with the Z-position history, \( Z \), we estimate the Z-positions of the hit points during probing. For the hit points which have positive Z-coordinates, the highest three values are averaged to determine \( Z_{\text{max}} \). Similarly, for hit points with negative Z-coordinates, the lowest three Z-values are averaged to compute \( Z_{\text{min}} \). If no hit points have positive Z-coordinates, $Z_{max}$ left empty. Similarly, $Z_{min}$ is empty if there are no hit points with negative Z-coordinates.


\subsubsection{Vertical body undulation control}
\label{Feedback controller}

\begin{figure}[h]
    \centering
    \includegraphics[width=9cm]{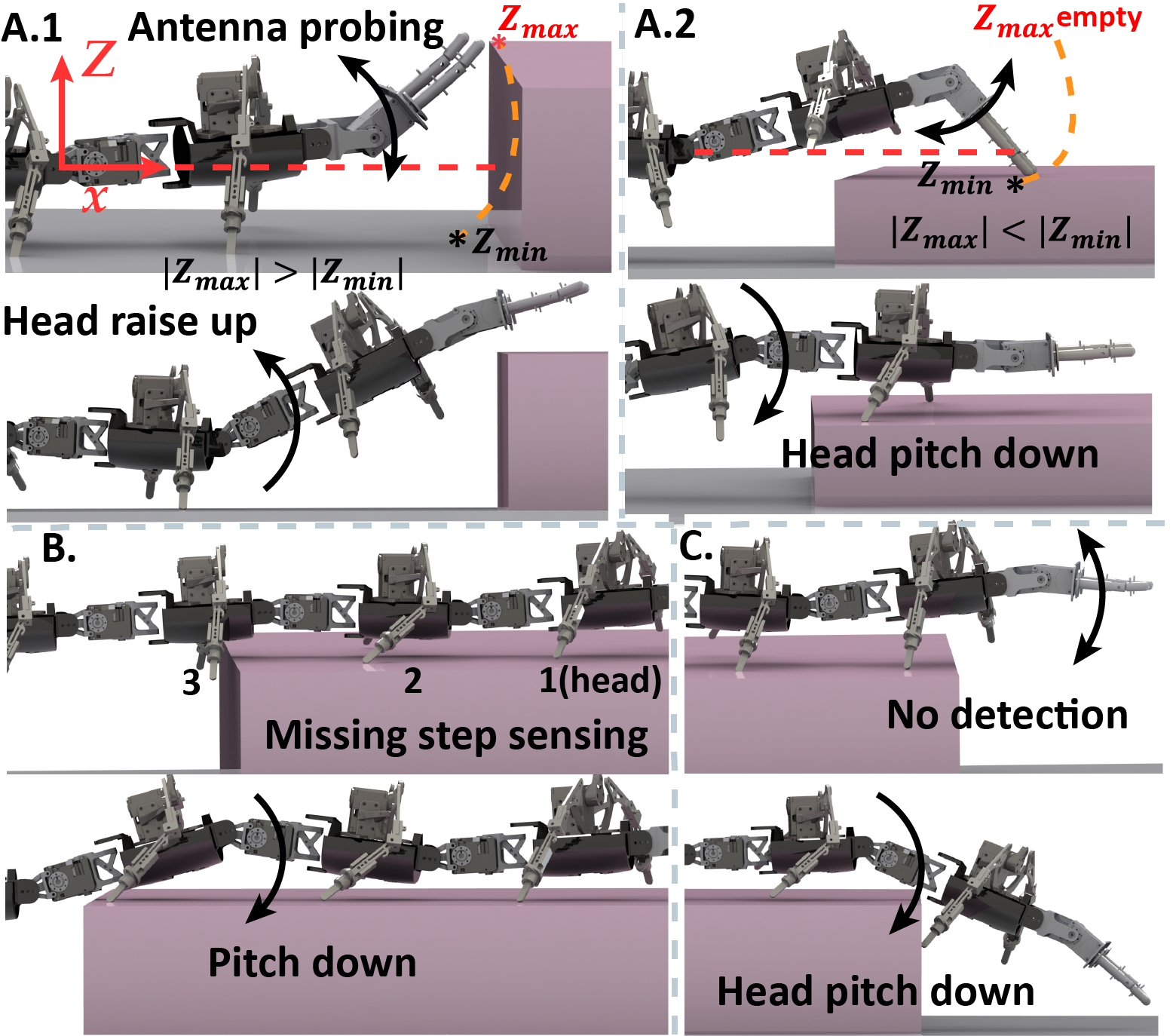}
    \caption{\textbf{Climbing controller working principle.} \textbf{A. Raise-up phase.} 1. The robot raises its head when the antenna detects an obstacle. 2. The robot pitches its head downward to position its legs on top of the obstacle once the antenna detects that the robot has cleared the obstacle. \textbf{B. Drag phase.} During the drag phase, the robot calculates the duty factor using data from ground-foot contact sensors to monitor the percentage of missed steps. A segment with a duty factor below the threshold is defined as floating, and the segment with the smallest index is identified as the one nearest to the edge. The closest vertical joint ahead of this segment then pitches downward, lifting the segment to the top of the obstacle. In this sketch, the 3rd segment has a duty factor below the threshold, so the second vertical joint is controlled to pitch down. \textbf{C. Special case.}  The sketch illustrates the special case where the antenna detects no obstacle. In this scenario, the robot is assumed to be beginning its descent, and a $2 \pi/9$ pitch-down motion for the head is hard coded to facilitate the process.}
    \label{fig:7}
\end{figure}

We divide the climbing process into two phases: ``raise up'' (Fig. \ref{fig:7}.A) and ``drag" (Fig. \ref{fig:7}.B). During the ``raise up" phase, the controller independently adjusts the vertical joint angle of the first (head) segment to position it quickly atop the obstacle. This motion is guided by a proportional (P) controller, which uses antenna measurements to estimate the obstacle's lowest and highest points. We choose to control only the head segment for the raise-up phase instead of additional segments for two reasons \cite{fu2020robotic,nilsson1998snake,yim2001climbing}. First, the motor's torque capability is sufficient to lift only one segment. Second, lifting two segments would cause severe instability. The first two segments and the antenna account for 45\% of the robot’s weight, making it easy for the center of mass to shift outside the base of support, resulting in unpredictable yawing and pitching motions. In the ``drag" phase, the head segment are fully positioned on the obstacle, pitching down the vertical joint near the obstacle edge speeds up the transition of the subsequent segment from floating to resting atop the obstacle.

We propose a head controller to regulate the vertical motion of the robot's head segment for both ``raise up" and ``drag" phase. This controller is inspired by the wall-following algorithm \cite{mongeau2015sensory,carelli2003corridor}. In wall-following, the robot is controlled to maintain a constant distance between its center and the wall. Similarly, in our approach, the head controller adjusts the pitch motion of the head segment to ensure the head joint maintains a constant distance from the contour of the obstacle. 

In the raise-up phase, the controller lifts the head segment above the obstacle (Fig. \ref{fig:7}.A1) and pitches it downward to place the legs on top of the obstacle (Fig \ref{fig:7}.A2). In the drag phase, the controller keeps the underside (belly) of the head segment closely aligned with the obstacle’s surface, allowing the legs to maintain ground contact. Consistent contact with the ground is essential, as the robot generates thrust by periodically engaging its legs with the surface \cite{chong2023multilegged,chong2023self}. This behavior is achieved using a proportional controller that leverages the history of $H$ and $Z$ values from the antenna.
\begin{figure}[h]
    \centering
    \includegraphics[width=8.5cm]{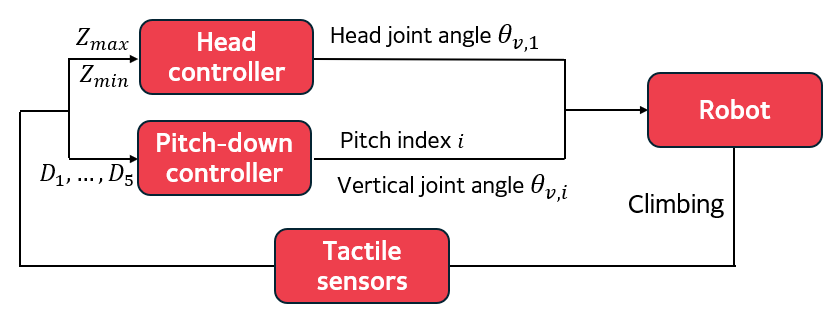}
    \caption{\textbf{Feedback control framework for climbing}. Two parallel controllers independently manage the pitch motion of the head segment and a subsequent segment. Using the Z-positions ($Z_{max}$ and $Z_{min}$) of the obstacle measured by the antenna, the head controller adjusts the head segment to closely follow the contour of the obstacle. Based on the duty factors ($D_1,...,D_5$) measured by foot contact sensors, the pitch-down controller identifies the appropriate floating segment nearest to the edge and pitches it down to accelerate the transition from floating to being positioned on top of the obstacle.}
    \label{fig:feedback}
\end{figure}

The controller first compares the absolute values of highest ($Z_{max}$) and lowest ($Z_{min}$) Z-position (Fig. \ref{fig:7}.A). If $|Z_{max}|$ is larger, then head segment is raised, with its vertical joint angle controlled by $K_p(Z_{max}+a)$. Here, $a$ is the distance between the head joint and the ground when the robot is in its initial state (Fig. \ref{fig:6}.A). Otherwise, the head segment pitches downward to keep a constant distance between the head joint and the obstacle contour, with the joint angle controlled by  $K_p(|Z_{min}|-a)$. 

In the special case where no ``hit" detection occurs (Fig. \ref{fig:7}.C), it indicates that the robot is descending back to the ground. In this scenario, the controller pitches the head segment downward by a constant value $\theta_0$. This hard coded downward motion ensures that the robot's legs land properly on the ground.

The vertical joint angle of the head segment is dynamically computed as:
\begin{equation}
     \theta_{v,1} = \left\{\begin{matrix}
 K_{p,1} (Z_{max}+a)& \text{for } Z_{max} \geq  |Z_{min}|, \\
 K_{p,2} (\left | Z_{min}\right |-a)& \text{for } Z_{max} < |Z_{min}|, \\
 \theta_0&  \text{for no hit detects}, \\
\end{matrix}\right.
\label{head controller}
\end{equation}
\noindent where $K_{p,1}$ and $K_{p,2}$ are proportional gains, and $\theta_0$ is a constant that pitches the head downward when no contact is detected. The constant $a$ accounts for the vertical offset between the head joint and the ground, as the local frame (Fig \ref{fig:6}) is defined at the head segment joint, located $a$ cm above the ground. The value of $a$ varies depending on the robot's dimensions.

During the drag phase, some segments are fully positioned on the obstacle, while subsequent segments and legs remain floating due to the level change (Fig. \ref{fig:7}.B top). This results in a slowdown as some legs lose ground contact, reducing the overall thrust.

To accelerate the transition over the obstacle, a localized pitch-down motion is applied to the vertical joint nearest to the obstacle's edge. The duty factor ($D$) of each segment is monitored using data from contact sensors, and segments with a duty factor below a threshold are identified as floating. Among these, the floating segment with the smallest index is considered closest to the edge. The vertical joint immediately ahead of this segment is then commanded to pitch downward (Fig. \ref{fig:7}.B). Note that the index starts from 2, as there is no vertical joint ahead of the first segment.

The pitch-down joint angle is defined as: \begin{equation} \theta_{v,i} = A_p \sin(4t) + \theta_p, \label{pitch-down controller} \end{equation} where $\theta_{v,i}$ is the vertical joint angle, $A_p$ is the amplitude of the periodic pitch motion, and $\theta_p$ is a negative constant that biases the motion downward.

In cases where $i = 1$, both the head controller and the pitch-down controller attempt to control the head simultaneously. To resolve this conflict, we assign higher priority to the pitch-down controller, ensuring that the head's motion is governed by it in such situations.

The overall feedback control framework is illustrated in Fig. \ref{fig:feedback}. Note that only the pitch motion of the head segment and the floating segment closest to the obstacle's edge are controlled by the controllers, while the limb stepping, horizontal body undulation, and the remaining vertical joint undulation follow the wave templates described in Section \ref{wave pattern}.

\subsubsection{Maximal vertical climbing capacity}
\begin{figure}[b]
    \centering
    \includegraphics[width=9cm]{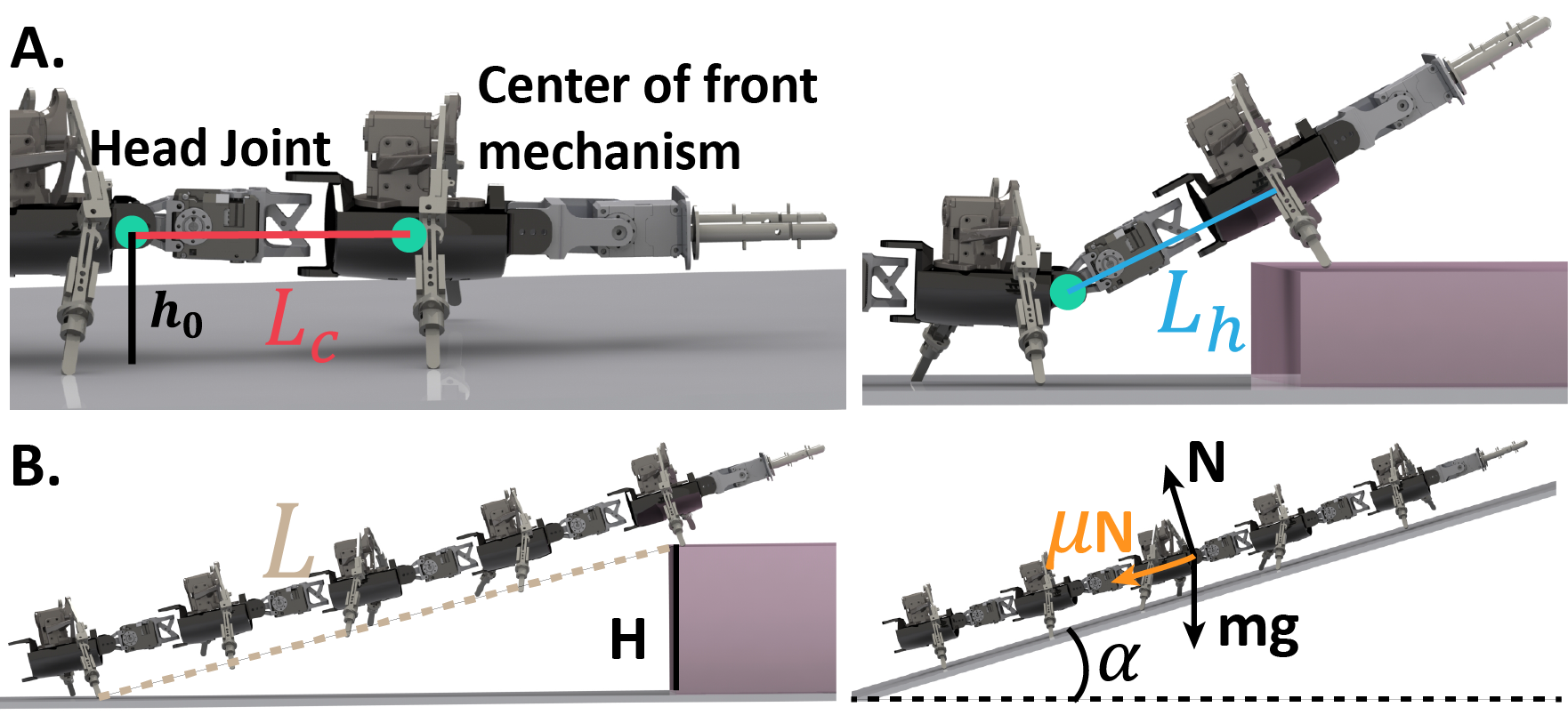}
    \caption{\textbf{Sketches for two types of bounds.} \textbf{A.} Illustrations of key distances: the distance between the head and center of mass of mechanism ahead of the head joint ($L_c$), and the distance between the head joint and the tip hook point ($L_h$). \textbf{B.} When the front legs land on the obstacle, gravity acts as resistance, slowing forward motion. This scenario can be approximated as movement on a slope, where the effective slope angle $\alpha$ is determined by the obstacle height $H$ and robot's length $L$.}
    \label{fig:climb capacity}
\end{figure}

In this section we will explore the upper bound of the vertical climbing capacity for our robot. Notably, for this analysis, we assume the obstacle has a box-like shape.

The first bound is determined by robot's dimension. In the raise-up phase, the head segment is lifted and positioned atop the obstacle. We define the pivot point as the point on the robot that makes contact with the top of the box. There are two possible pivot points: the belly or the leg.

We first consider the case where the pivot point is on the belly. In this scenario, the maximum theoretical height the robot can hook onto is determined by $L_c$, the distance between the head joint and the center of mass of mechanism ahead of the head joint (Fig. \ref{fig:climb capacity}). Specifically, we consider a pivot point to be stable if it lies on $L_c$, as pivot points along this line ensure that the center of mass of the front mechanism enters the obstacle's region.


Notably, the robot's legs can also function as pivot points \cite{song2022gait,eich2008versatile}, providing additional anchoring to enhance the head-hooking process (Fig.\ref{fig:climb capacity}.A). The corresponding maximum box height is denoted as $L_h$, which represents the distance between the head joint and the farthest front legs. This anchoring increases the climbing height limit, which can be expressed as:
\begin{equation*}
    b_1: H_{max} < h_{0} + \text{max}\{L_c, L_h\},
\end{equation*} 
where $h_{0}$ is the maximum distance between the head joint and the ground over gait cycle. Note that, parameter $h_{0}$ functions as an offset that raises the belly above the ground.

The second bound arises from the mechanics of the robot. During the drag phase, the robot naturally forms a slope along its backbone due to the level difference between the head and tail segments. In this configuration, gravity acts as a resistive force, opposing forward motion. As Fig.\ref{fig:climb capacity}.B shows, the robot's motion can be conceptualized as movement on an inclined plane. When the effective ``slope" angle $\alpha$—determined by the robot’s geometry and the obstacle height—exceeds a certain threshold, the robot's forward movement becomes significantly hindered, and it may barely be able to move forward. Here, we define the threshold as 80\% of the friction angle and the denote the corresponding obstacle height as $H'$. Then the second bound $H'$ is computed as $\sin (0.8\tan^{-1}(\mu)) L$, where $L$ is length between legs on the head and tail segments. 

Finally, the combination of two types of bounds is expressed as:
\begin{align}
     b_1:& H_{max} < h_{0} +\text{max}\{L_c,L_h\}.\\
     b_2: &H_{max} < \sin (0.8\tan^{-1}(\mu)) L
     \label{eq:constraint}
\end{align}

Given the robot's dimension and wave pattern parameters in this paper, we have $h_{0}= 7$ cm, $L_c = 15$ cm, $\mu \approx 0.5$, $L_h = 17$ cm, $L = 95$ cm and $H' = 34$ cm. Therefore, the estimated obstacle height limit is approximately 24 cm.

\subsection{Experiment results}
\begin{figure}
    \centering
    \includegraphics[width=9cm]{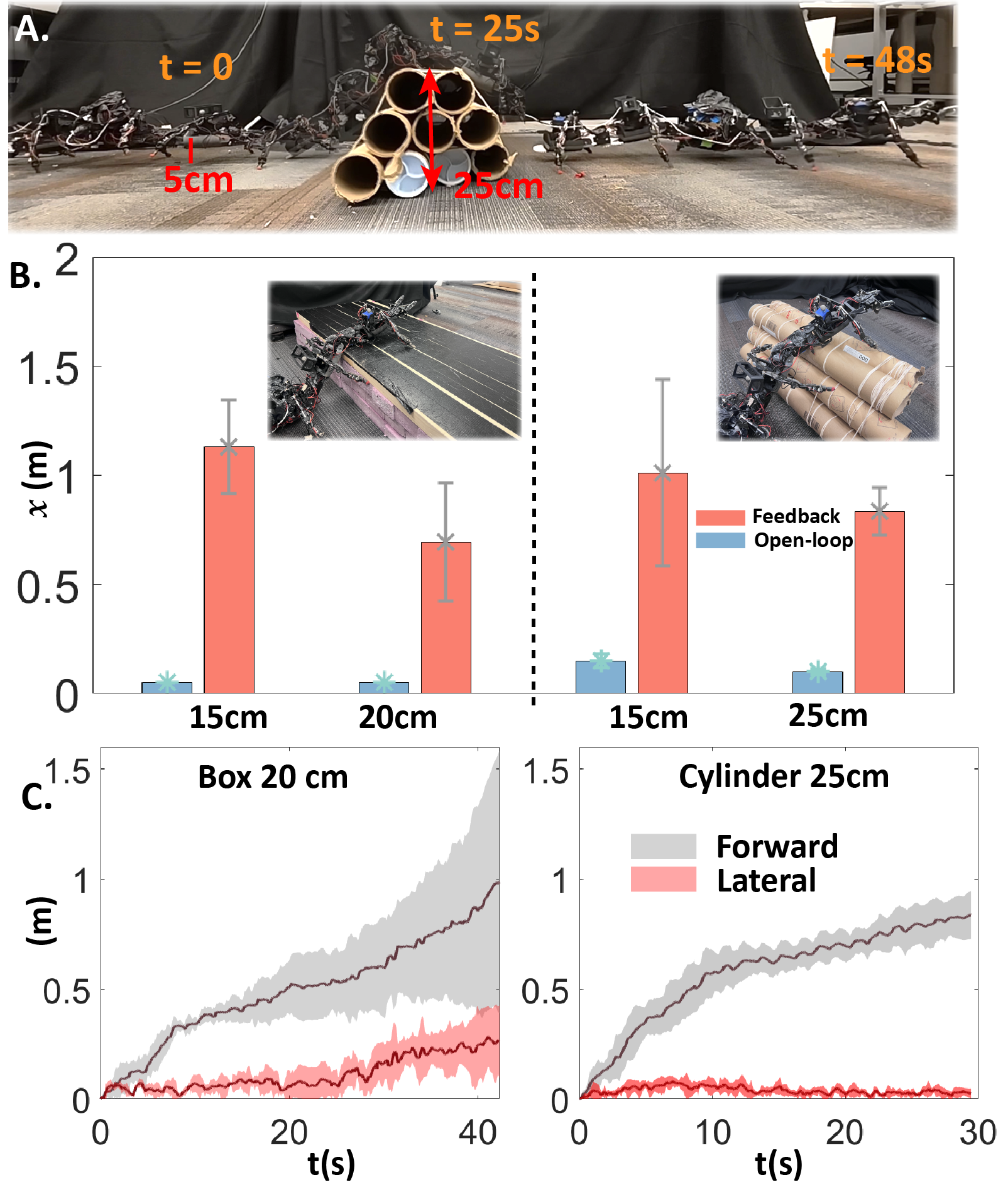}
    \caption{\textbf{Locomotion performance comparison between open-loop and feedback controllers across two obstacle types.} \textbf{A.} Snapshots of the robot traversing a 25 cm bundled-cylinder (five times the robot's height) using the feedback controller. \textbf{B.} The $x$ indicates the robot's forward displacement. The left half of the graph corresponds to the box-shaped obstacle, and the right half corresponds to the bundled-cylinder obstacle. We conducted 10 trials for each experiment. 
    \textbf{C.} This plot illustrates the forward and lateral displacement history of the robot as it traverses a 20 cm box-shaped obstacle and a 25 cm bundled-cylinder obstacle.}
    \label{fig:9}
\end{figure}
We evaluated our climbing feedback controller in both laboratory-based and outdoor environments. In these experiments, the limb stepping wave amplitude (\( \Theta_{\text{leg}} \)) was set to \( \pi/6 \), the horizontal body wave amplitude (\( \Theta_{\text{body}} \)) to \( \pi/18 \), and the vertical wave amplitude (\( A_v \)) to \( \pi/9 \). The wavenumber of the three waves is set to 1.5. These parameters were chosen to optimize the robot's speed while minimizing yaw motion during climbing, based on findings from preliminary work~\cite{chong2023multilegged,chong2023self,he2024probabilistic} and empirical experiments. The frequency of pitch motion adjustments made by the controllers is every quarter cycle of motion. The $K_{p,1}$,$K_{p,2}$, $a$ and $\theta_0$ in Eq. \ref{head controller} are set as 2, 1, 6 and $2\pi/9$ in all experiments. $A_p$ and $\theta_p$ in Eq. \ref{pitch-down controller} are set as $\pi/12$ and $\pi/6$ respectively. These parameters are tuned empirically to be most effective for our experiments. 

\subsubsection{Single obstacle}
For the laboratory experiments, we tested the robot's ability to navigate box-shaped obstacles (120 cm length $\times$ 45 cm width, Fig. \ref{fig:9}.b) with heights of 15 cm and 20 cm, as well as bundled-cylinder obstacles (Fig. \ref{fig:9}.b) with heights of 15 cm and 25 cm. The second obstacle is a trapezoidal prism constructed by stacking layers of cardboard tubes with progressively decreasing widths. The base layer consisted of four tubes, each 9 cm in diameter and approximately 70 cm in length. The cylindrical shape was chosen to evaluate the controller's robustness when navigating rapidly changing curvatures. We conducted 10 trials of test on each obstacle.

In each trial, the robot was positioned with its antenna 5 cm in front of the obstacle. For the experiments on box-shaped obstacles, the robot was operated for 14 motion cycles (a total of 42 seconds). For the experiments on bundled-cylinder obstacles, the robot was operated for 10 motion cycles (a total of 30 seconds). The total displacement of those experiments are plotted in Fig. \ref{fig:9}.B. According to the results, the robot successfully climbs a 20 cm box-shaped obstacle, which is just below the 24 cm theoretical limit derived in the previous section. Additionally, we observe that the climbing limit increases when the robot is tested on the bundled-cylinder obstacle. This is because the obstacle is not entirely vertical; its slope (around $60^{\circ}$) raises $h_0$ in Eq.\ref{eq:constraint}, thus raises the climbing limit. 
\begin{figure}
    \centering
    \includegraphics[width=9cm]{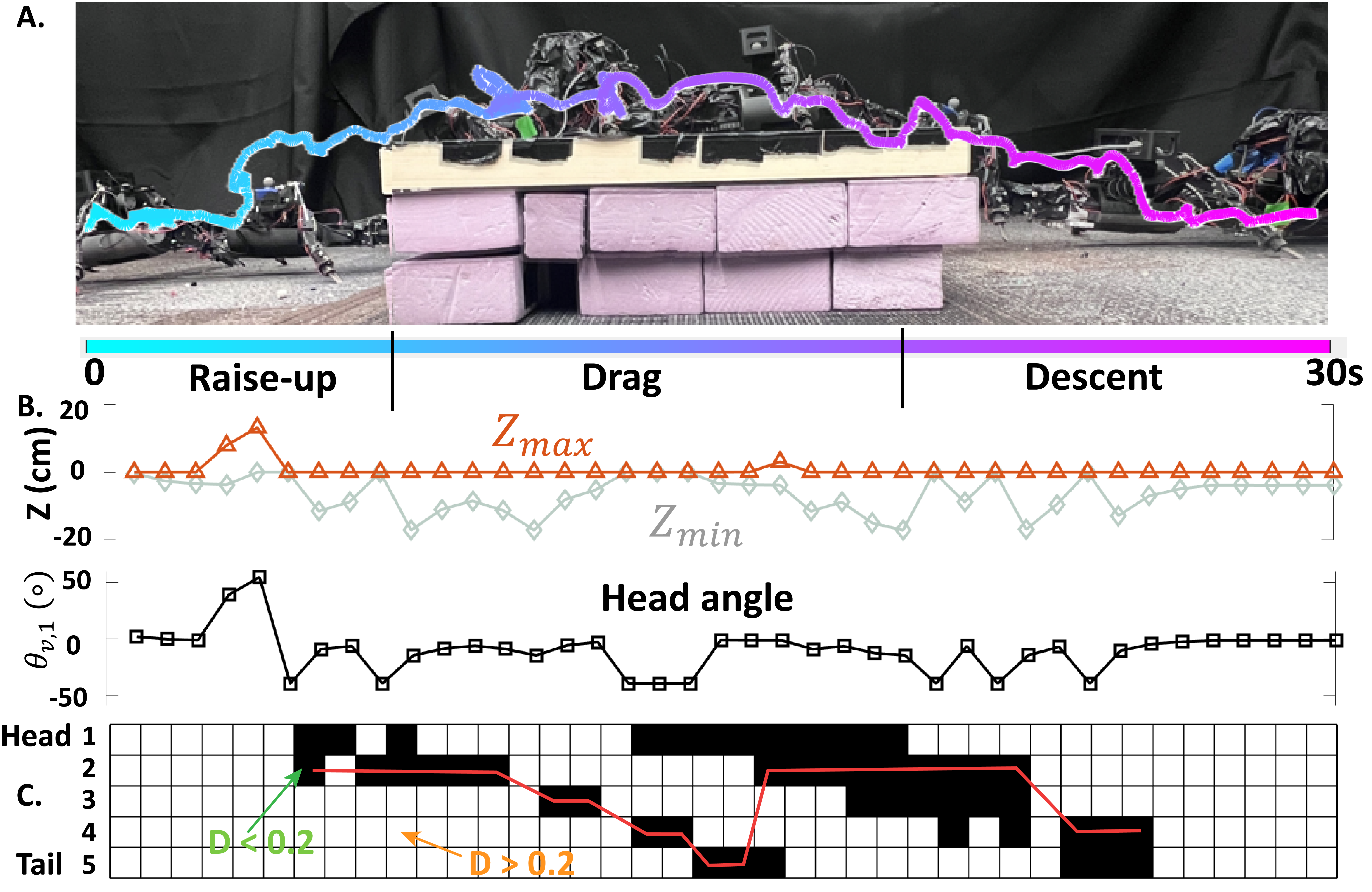}
    \caption{\textbf{Robot displacement, obstacle height estimation, and segment duty factor tracking history.} \textbf{A.} The geometric center history of the robot during a single trial on a 15 cm box-shaped obstacle. \textbf{B.} (Top) The history of $Z_{max}$ and $Z_{min}$ estimated by the antenna. If $Z_{max}$ or $Z_{min}$ is empty, we set its value to 0 to simplify data visualization. (Bottom) The history of the head joint angle determined by the Z-position data. \textbf{C.} The duty factor history for each segment. Black indicates duty factors below the threshold ($D<0.2$), and the red curve connects the floating segment closest to the obstacle edge. The index starts from 2 because there is no vertical joint ahead of the first segment.}
    \label{fig:8}
\end{figure}

Additionally, we provide a plot showcasing detailed antenna and contact sensor data from a trial in which the robot successfully climbed a 15 cm box-shaped obstacle. In this trial, the robot was positioned with its antenna 5 cm in front of the obstacle and ran for 10 motion cycles. This visualization is intended to help readers better understand the working principles of our controller.

In Fig. \ref{fig:8}, \textbf{A} shows the trajectory of the robot's center of geometry. The top of \textbf{B} displays the history of \( Z_{\text{max}} \) and \( Z_{\text{min}} \), as measured by the antenna, while the bottom of \textbf{B} illustrates the history of the head segment's vertical joint angle, adjusted by the controller based on the measured \( Z \)-position data. \textbf{C} records the duty factor history for each segment, with segments shown in black when the duty factor falls below the threshold of 0.2. The red curves connect the floating segment closet to the obstacle's edge. Since the indices of floating segments start at 2, the black grids are not connected for the first segment.

Recall that the head controller raises the head segment when the antenna detects an obstacle and subsequently lowers it to ensure stable leg contact with the ground. As shown in Fig. \ref{fig:8}.B, during the raise-up phase, the antenna first detects the obstacle, prompting the head joint angle to gradually increase. Once the antenna determines that the area ahead of the robot is clear, the head pitches down to place the legs securely on the ground.

Following this, the robot transitions into the drag phase as the first segment is fully positioned on top of the obstacle. In this phase, the head controller regulates the vertical motion of the head segment to maintain a consistent distance between the robot's belly and the contour of the obstacle.

When the head reaches the far edge of the obstacle and begins to leave its surface, the robot enters the descent phase. During this phase, the antenna may either detect no obstacle or measure a Z-position lower than the head segment joint. As expected, the head controller pitches the head downward to facilitate a smooth descent.

In addition, the pitch-down motion controller adjusts the floating segment closest to the obstacle's edge, pitching it downward to expedite its transition from floating to being positioned on top of the obstacle. As shown in Fig. \ref{fig:8}.C, during the drag phase, the index of the floating segments shifts sequentially from 2 to 5, indicating that these segments become closest to the edge in succession. The pitch-down controller dynamically adjusts the nearest vertical joint ahead of these segments, facilitating a faster transition.
\begin{figure}[h]
    \centering
    \includegraphics[width=8.5cm]{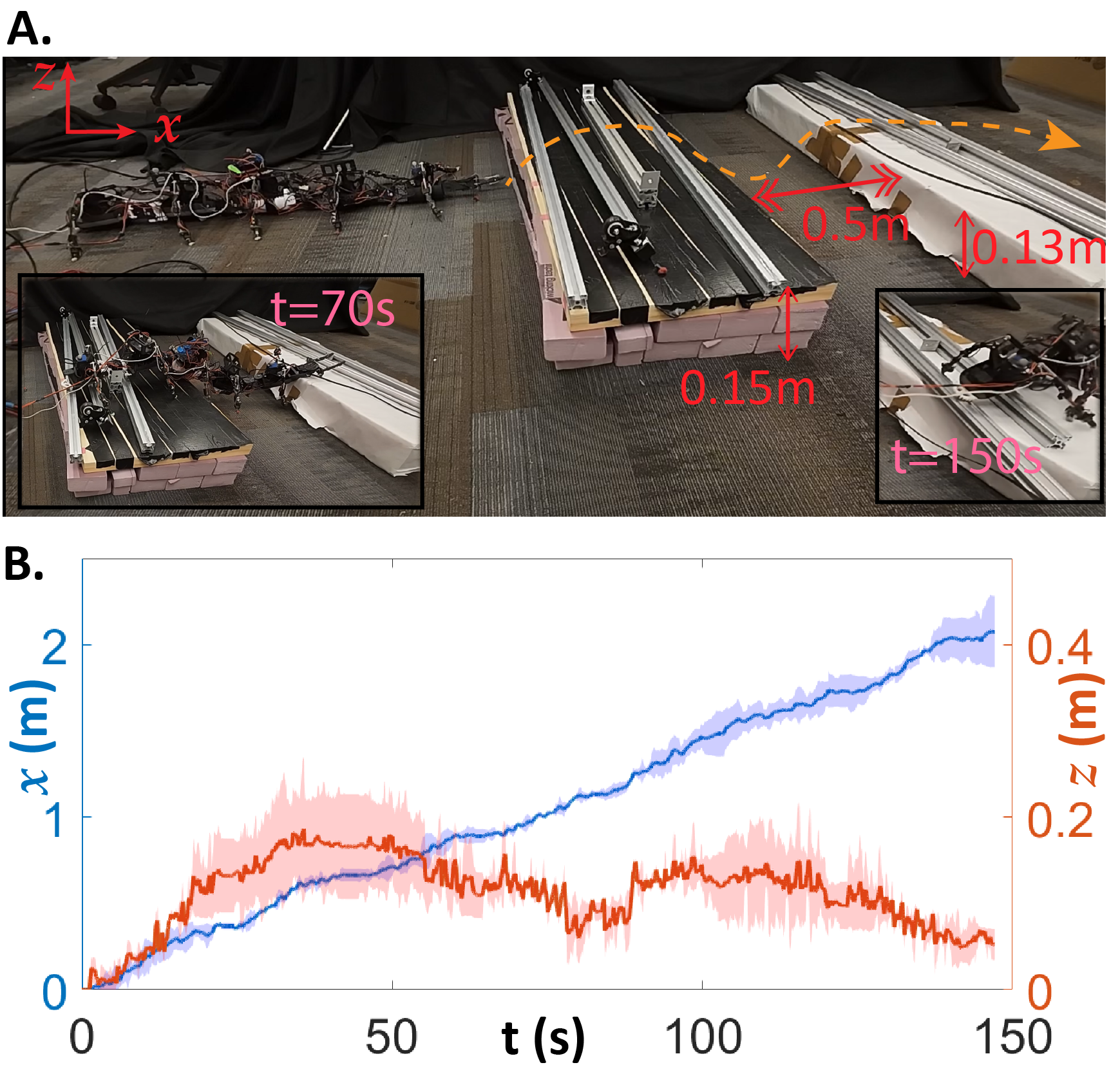}
    \caption{\textbf{Traversal of two consecutive obstacles} A. Snapshots showing the robot traversing two obstacles (0.15 m and 0.13 m in height, spaced 0.5 m apart) within 150 s. B. Forward ($x$) and vertical ($z$) displacement across five trials. Shaded regions represent the standard deviation.}
    \label{fig:two obstacle}
\end{figure}
\subsubsection{Multiple obstacles in close succession}
To evaluate the robustness of the feedback controller in handling multiple obstacles, we tested the robot on two consecutive obstacles (0.15 m and 0.13 m in height), spaced 0.5 m apart (Fig. \ref{fig:two obstacle}.A), ensuring that the tail remained on the first obstacle as the head ascended the second. Both obstacles were covered with metal beams to increase the complexity of the task. The robot successfully traversed both obstacles in all five trials. Tracking results for forward and vertical displacement are shown in Fig.\ref{fig:1}.B.
\subsubsection{Shifting obstacles}
\begin{figure}
    \centering
    \includegraphics[width=8.5cm]{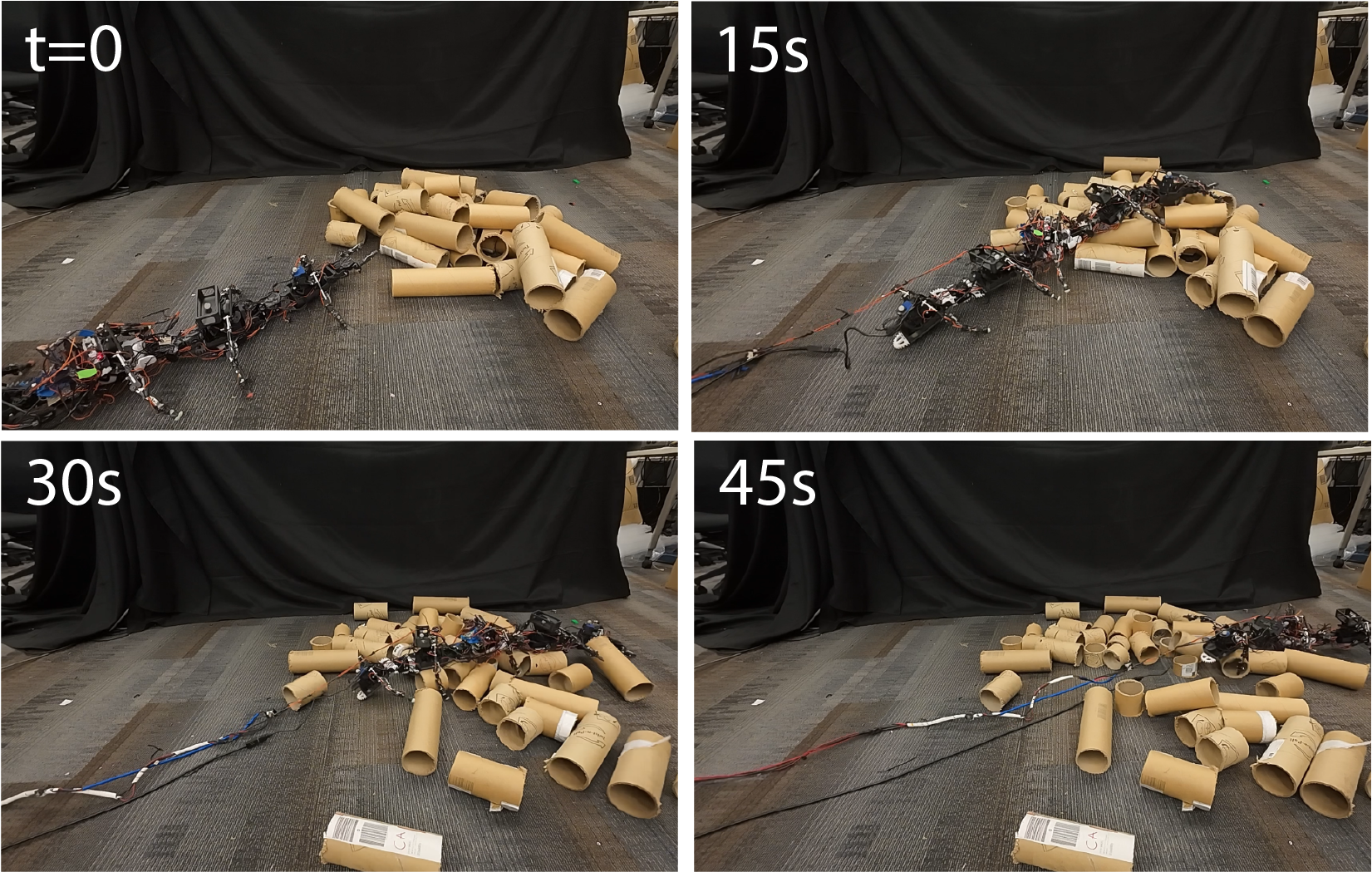}
    \caption{\textbf{Robot navigating unstable obstacles.} Snapshots show the robot traversing a pile of unstable, shifting obstacles — 9 cm diameter cylinders of varying lengths (5 – 15 cm), covering a 60 cm × 60 cm area and reaching up to 20 cm in height.}
    \label{fig:shifting obstacle}
\end{figure}

We also tested the robot on a pile of shifting cylinders (Fig. \ref{fig:shifting obstacle}). The pile covered a 60 cm × 60 cm area and reached a height of up to 20 cm, consisting of 9 cm diameter cylinders with varying lengths (5 – 15 cm). In all five trials, the robot successfully reached the peak and exited the pile. Fig. \ref{fig:shifting obstacle} shows snapshots from a representative trial in which the obstacles collapsed and shifted during traversal.
\subsubsection{Outdoor test}
\begin{figure*}
    \centering
    \includegraphics[width=18cm]{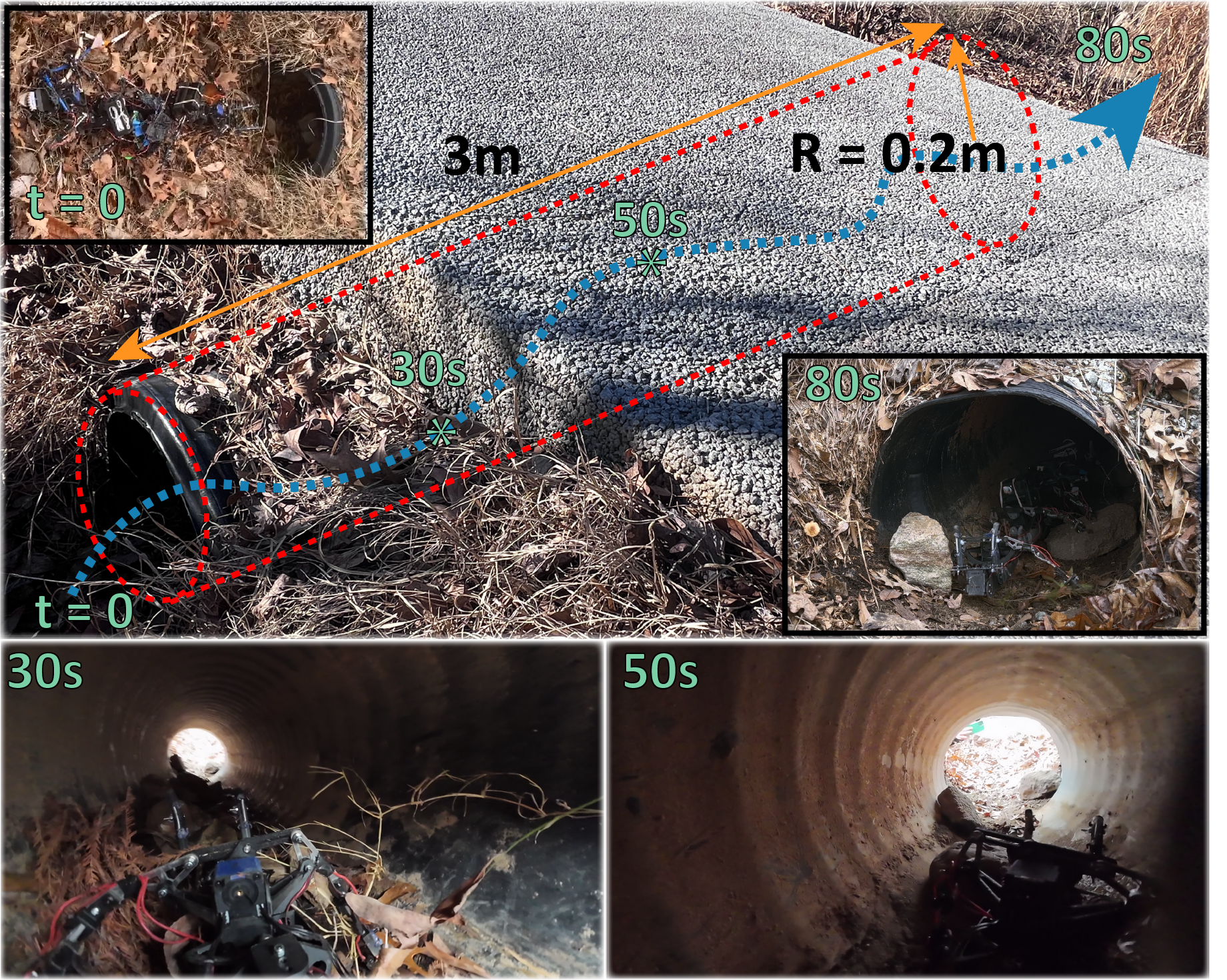}
    \caption{\textbf{Robot traverses a pipe.} The robot completed an ``inspection" task in 80 seconds within a cylindrical pipe measuring 0.2 m in radius and 3 m in length. The pipe contained randomly placed robot-sized rocks, requiring the robot's climbing capability, while leaves and pine straw introduced entanglement challenges, increasing task complexity. A camera mounted on the robot captured footage during the traversal.}
    \label{fig:outdoor}
\end{figure*}

For the outdoor tests, we evaluated the robot's climbing ability in confined spaces containing robot-sized rocks, mud, and leaves. The robot successfully climbed obstacles up to four times its height (20 cm) in these challenging environments. Fig. \ref{fig:outdoor} shows the robot traversing a 0.2 m radius, 3 m long pipe in 80 seconds. The pipe contained randomly placed robot-sized rocks, requiring the robot to utilize its climbing capabilities, while leaves and pine straw introduced entanglement challenges, further increasing task complexity. An onboard sports camera mounted on the robot captured footage for inspection. A compilation of outdoor tests is available in the SI video.

\section{Conclusion} 
\label{sec:conclusion}
In this work, we identified an alternative approach for perceiving and responding to the environment during legged locomotion, contrasting with conventional vision-based control frameworks. Specifically, we introduced a tactile-based antenna design that enabled an elongate many-legged robot to sense short-range contact information and reconstruct rough geometric features of obstacles. By integrating antenna sensing with another low-bandwidth foot contact tactile sensor, the mechanically intelligent many-legged robot demonstrates predictable climbing performance in complex environments. 

Specifically, we first investigated the robot's maximum climbing capability under open-loop control, achieved by coordinating limb stepping with horizontal and vertical body undulation wave patterns. Building on this foundation, we developed a control framework that integrates antenna data and binary foot-ground contact information to dynamically plan vertical body undulation, enabling effective climbing. Laboratory and outdoor experiments demonstrated the robot's ability to climb obstacles up to five times its center height and navigate challenging terrains, including obstacles with rapidly changing curvatures and those covered with flowable, robot-sized random items. The robustness of this climbing controller was further validated in complex outdoor environments. For example, the robot successfully traversed a 0.2 m radius, 3 m long pipe filled with robot-sized rocks, leaves, and pine straw.

These findings underscore the value of integrating tactile sensory systems and control frameworks address the challenges of 3D motion planning in many-legged robots. Given its 3D motion capability in confined spaces, the robot also shows strong potential for pipe inspection \cite{ismail2012development,verma2022review,nayak2014design}. 

We used GPT to refine our writing.
\section{Limitations}
In this work, we focused on how the antenna enables climbing for a many-legged robot. With its ability to provide short-range estimations of obstacle geometry, the antenna's potential extends beyond climbing to include functionalities such as obstacle avoidance \cite{gibson2018evaluation,buchs2014obstacle,lobo2019route}, which could further enhance the robot's versatility in complex environments. Exploring these additional capabilities is a key area for future research.

We observed that the robot successfully climbs obstacles when no strict height limit is imposed. However, the current controller struggles to differentiate between ``ceilings" and climbable obstacles, leading to failures in specific scenarios. To address this limitation, we plan to develop a filtering mechanism in the control framework to correctly interpret such cases, ensuring robust and reliable performance across diverse terrains.

The current antenna design (Fig. \ref{fig:3}) includes a compliant section near its tip. Since the base of the antenna is rigid, it occasionally experiences jamming when interacting with obstacles. To mitigate this issue, we propose replacing the current design with a TPU-printed antenna. This modification would enhance flexibility, reduce the impact of environmental interactions, and improve overall performance.

In our feedback control framework, a linear controller adjusts the head segment's pitch angle based on $Z_{max}$ and $Z_{min}$, while the pitch-down motion of the other vertical joints is predefined based on missing step information derived from the duty factors [$D_1,…,D_5$]. However, the optimal mapping from sensory inputs to joint angles is likely more complex and nonlinear. In future work, we plan to implement reinforcement learning to train a learning-based policy, where neural networks will provide a more sophisticated nonlinear mapping. We anticipate that this RL-based approach will further enhance the robot's climbing speed.

In our analysis of climbing capacity, we found that adhesion can aid in climbing performance. In future work, we plan to explore adhesion-based mechanisms for the robot’s feet or belly to further enhance its climbing ability.

\section{Acknowledgment}
We thank Daniel Soto and Esteban Flores for the useful discussion on the robot mechanical design. We are grateful for funding from NSF STTR AWD-2335553, Army Research Office Grant AWD-001950, Georgia Research Alliance (GRA) AWD-005494, GA AIM AWD-004173 and the Dunn Family Professorship.

\bibliographystyle{plainnat}
\bibliography{references}

\begin{thebibliography}{57}
\providecommand{\natexlab}[1]{#1}
\providecommand{\url}[1]{\texttt{#1}}
\expandafter\ifx\csname urlstyle\endcsname\relax
  \providecommand{\doi}[1]{doi: #1}\else
  \providecommand{\doi}{doi: \begingroup \urlstyle{rm}\Url}\fi

\bibitem[Aoi et~al.(2022)Aoi, Tomatsu, Yabuuchi, Morozumi, Okamoto, Fujiki, Senda, and Tsuchiya]{aoi2022advanced}
Shinya Aoi, Ryoe Tomatsu, Yuki Yabuuchi, Daiki Morozumi, Kota Okamoto, Soichiro Fujiki, Kei Senda, and Kazuo Tsuchiya.
\newblock Advanced turning maneuver of a many-legged robot using pitchfork bifurcation.
\newblock \emph{IEEE Transactions on Robotics}, 38\penalty0 (5):\penalty0 3015--3026, 2022.

\bibitem[Berman et~al.(2024)Berman, Hsiao, Root, Choi, Ilyn, Xu, Stein, Cutkosky, DeSimone, and Bao]{berman2024additively}
Arielle Berman, Kaiwen Hsiao, Samuel~E Root, Hojung Choi, Daniel Ilyn, Chengyi Xu, Emily Stein, Mark Cutkosky, Joseph~M DeSimone, and Zhenan Bao.
\newblock Additively manufactured micro-lattice dielectrics for multiaxial capacitive sensors.
\newblock \emph{Science Advances}, 10\penalty0 (40):\penalty0 eadq8866, 2024.

\bibitem[Bretl(2006)]{bretl2006motion}
Timothy Bretl.
\newblock Motion planning of multi-limbed robots subject to equilibrium constraints: The free-climbing robot problem.
\newblock \emph{The International Journal of Robotics Research}, 25\penalty0 (4):\penalty0 317--342, 2006.

\bibitem[Buchs et~al.(2014)Buchs, Maidenbaum, and Amedi]{buchs2014obstacle}
Galit Buchs, Shachar Maidenbaum, and Amir Amedi.
\newblock Obstacle identification and avoidance using the ‘eyecane’: a tactile sensory substitution device for blind individuals.
\newblock In \emph{Haptics: Neuroscience, Devices, Modeling, and Applications: 9th International Conference, EuroHaptics 2014, Versailles, France, June 24-26, 2014, Proceedings, Part II 9}, pages 96--103. Springer, 2014.

\bibitem[Carelli and Freire(2003)]{carelli2003corridor}
Ricardo Carelli and Eduardo~Oliveira Freire.
\newblock Corridor navigation and wall-following stable control for sonar-based mobile robots.
\newblock \emph{Robotics and Autonomous Systems}, 45\penalty0 (3-4):\penalty0 235--247, 2003.

\bibitem[Charlebois et~al.(1996)Charlebois, Gupta, and Payandeh]{charlebois1996curvature}
Mark Charlebois, Kamal Gupta, and Shahram Payandeh.
\newblock Curvature based shape estimation using tactile sensing.
\newblock In \emph{Proceedings of IEEE International Conference on Robotics and Automation}, volume~4, pages 3502--3507. IEEE, 1996.

\bibitem[Chen et~al.(2024)Chen, Newdick, Di, Bosio, Ongole, Lap{\^o}tre, Pavone, and Cutkosky]{chen2024locomotion}
Tony~G Chen, Stephanie Newdick, Julia Di, Carlo Bosio, Nitin Ongole, Mathieu Lap{\^o}tre, Marco Pavone, and Mark~R Cutkosky.
\newblock Locomotion as manipulation with reachbot.
\newblock \emph{Science Robotics}, 9\penalty0 (89):\penalty0 eadi9762, 2024.

\bibitem[Cheng et~al.(2024)Cheng, Shi, Agarwal, and Pathak]{cheng2024extreme}
Xuxin Cheng, Kexin Shi, Ananye Agarwal, and Deepak Pathak.
\newblock Extreme parkour with legged robots.
\newblock In \emph{2024 IEEE International Conference on Robotics and Automation (ICRA)}, pages 11443--11450. IEEE, 2024.

\bibitem[Chong et~al.(2022)Chong, Aydin, Rieser, Sartoretti, Wang, Whitman, Kaba, Aydin, McFarland, Cruz, et~al.]{chong2022general}
Baxi Chong, Yasemin~O Aydin, Jennifer~M Rieser, Guillaume Sartoretti, Tianyu Wang, Julian Whitman, Abdul Kaba, Enes Aydin, Ciera McFarland, Kelimar~Diaz Cruz, et~al.
\newblock A general locomotion control framework for multi-legged locomotors.
\newblock \emph{Bioinspiration \& Biomimetics}, 17\penalty0 (4):\penalty0 046015, 2022.

\bibitem[Chong et~al.(2023{\natexlab{a}})Chong, He, Li, Erickson, Diaz, Wang, Soto, and Goldman]{chong2023self}
Baxi Chong, Juntao He, Shengkai Li, Eva Erickson, Kelimar Diaz, Tianyu Wang, Daniel Soto, and Daniel~I Goldman.
\newblock Self-propulsion via slipping: Frictional swimming in multilegged locomotors.
\newblock \emph{Proceedings of the National Academy of Sciences}, 120\penalty0 (11):\penalty0 e2213698120, 2023{\natexlab{a}}.

\bibitem[Chong et~al.(2023{\natexlab{b}})Chong, He, Soto, Wang, Irvine, Blekherman, and Goldman]{chong2023multilegged}
Baxi Chong, Juntao He, Daniel Soto, Tianyu Wang, Daniel Irvine, Grigoriy Blekherman, and Daniel~I Goldman.
\newblock Multilegged matter transport: A framework for locomotion on noisy landscapes.
\newblock \emph{Science}, 380\penalty0 (6644):\penalty0 509--515, 2023{\natexlab{b}}.

\bibitem[Eich et~al.(2008)Eich, Grimminger, and Kirchner]{eich2008versatile}
Markus Eich, Felix Grimminger, and Frank Kirchner.
\newblock A versatile stair-climbing robot for search and rescue applications.
\newblock In \emph{2008 IEEE international workshop on safety, security and rescue robotics}, pages 35--40. IEEE, 2008.

\bibitem[Fearing and Binford(1988)]{fearing1988using}
Ronald~S Fearing and Thomas~O Binford.
\newblock Using a cylindrical tactile sensor for determining curvature.
\newblock In \emph{Proceedings. 1988 IEEE International Conference on Robotics and Automation}, pages 765--771. IEEE, 1988.

\bibitem[Fu and Li(2020)]{fu2020robotic}
Qiyuan Fu and Chen Li.
\newblock Robotic modelling of snake traversing large, smooth obstacles reveals stability benefits of body compliance.
\newblock \emph{Royal Society open science}, 7\penalty0 (2):\penalty0 191192, 2020.

\bibitem[Fu and Li(2023)]{fu2023contact}
Qiyuan Fu and Chen Li.
\newblock Contact feedback helps snake robots propel against uneven terrain using vertical bending.
\newblock \emph{Bioinspiration \& Biomimetics}, 18\penalty0 (5):\penalty0 056002, 2023.

\bibitem[Gao and Kikuchi(2004)]{gao2004study}
Xueshan Gao and Koki Kikuchi.
\newblock Study on a kind of wall cleaning robot.
\newblock In \emph{2004 IEEE International Conference on Robotics and Biomimetics}, pages 391--394. IEEE, 2004.

\bibitem[Gao et~al.(2008)Gao, Jiang, Gao, Xu, Wang, and Pan]{gao2008boiler}
Xueshan Gao, Zhihong Jiang, Junyao Gao, Dianguo Xu, Yan Wang, and HuanHuan Pan.
\newblock Boiler maintenance robot with multi-operational schema.
\newblock In \emph{2008 IEEE International Conference on Mechatronics and Automation}, pages 610--615. IEEE, 2008.

\bibitem[Gibson et~al.(2018)Gibson, Webb, and Stirling]{gibson2018evaluation}
Alison Gibson, Andrea Webb, and Leia Stirling.
\newblock Evaluation of a visual--tactile multimodal display for surface obstacle avoidance during walking.
\newblock \emph{IEEE Transactions on Human-Machine Systems}, 48\penalty0 (6):\penalty0 604--613, 2018.

\bibitem[Haynes et~al.(2009)Haynes, Khripin, Lynch, Amory, Saunders, Rizzi, and Koditschek]{haynes2009rapid}
G~Clark Haynes, Alex Khripin, Goran Lynch, Jonathan Amory, Aaron Saunders, Alfred~A Rizzi, and Daniel~E Koditschek.
\newblock Rapid pole climbing with a quadrupedal robot.
\newblock In \emph{2009 IEEE international conference on robotics and automation}, pages 2767--2772. IEEE, 2009.

\bibitem[He et~al.(2024{\natexlab{a}})He, Chong, Lin, Xu, Bagheri, Flores, and Goldman]{he2024probabilistic}
Juntao He, Baxi Chong, Jianfeng Lin, Zhaochen Xu, Hosain Bagheri, Esteban Flores, and Daniel~I Goldman.
\newblock Probabilistic approach to feedback control enhances multi-legged locomotion on rugged landscapes.
\newblock \emph{arXiv preprint arXiv:2411.07183}, 2024{\natexlab{a}}.

\bibitem[He et~al.(2024{\natexlab{b}})He, Chong, Xu, Flores, Soto, and Goldman]{he2024tactile}
Juntao He, Baxi Chong, Zhaochen Xu, Esteban Flores, Daniel Soto, and Daniel Goldman.
\newblock Tactile feedback enhances multi-legged locomotion on rugged landscapes.
\newblock \emph{Bulletin of the American Physical Society}, 2024{\natexlab{b}}.

\bibitem[He et~al.(2024{\natexlab{c}})He, Chong, Xu, Ha, and Goldman]{he2024learning}
Juntao He, Baxi Chong, Zhaochen Xu, Sehoon Ha, and Daniel~I Goldman.
\newblock Learning to enhance multi-legged robot on rugged landscapes.
\newblock \emph{arXiv preprint arXiv:2409.09473}, 2024{\natexlab{c}}.

\bibitem[Hoeller et~al.(2024)Hoeller, Rudin, Sako, and Hutter]{hoeller2024anymal}
David Hoeller, Nikita Rudin, Dhionis Sako, and Marco Hutter.
\newblock Anymal parkour: Learning agile navigation for quadrupedal robots.
\newblock \emph{Science Robotics}, 9\penalty0 (88):\penalty0 eadi7566, 2024.

\bibitem[Hong et~al.(2022)Hong, Um, Park, and Park]{hong2022agile}
Seungwoo Hong, Yong Um, Jaejun Park, and Hae-Won Park.
\newblock Agile and versatile climbing on ferromagnetic surfaces with a quadrupedal robot.
\newblock \emph{Science Robotics}, 7\penalty0 (73):\penalty0 eadd1017, 2022.

\bibitem[Ismail et~al.(2012)Ismail, Anuar, Sahari, Baharuddin, Fairuz, Jalal, and Saad]{ismail2012development}
Iszmir~Nazmi Ismail, Adzly Anuar, Khairul Salleh~Mohamed Sahari, Mohd~Zafri Baharuddin, Muhammad Fairuz, Abd Jalal, and Juniza~Md Saad.
\newblock Development of in-pipe inspection robot: A review.
\newblock In \emph{2012 IEEE Conference on Sustainable Utilization and Development in Engineering and Technology (STUDENT)}, pages 310--315. IEEE, 2012.

\bibitem[Lee et~al.(2012)Lee, Wu, Kim, Kim, and Seo]{lee2012combot}
Giuk Lee, Geeyun Wu, Sun~Ho Kim, Jongwon Kim, and TaeWon Seo.
\newblock Combot: Compliant climbing robotic platform with transitioning capability and payload capacity.
\newblock In \emph{2012 IEEE International Conference on Robotics and Automation}, pages 2737--2742. IEEE, 2012.

\bibitem[Lee et~al.(2008)Lee, Sponberg, Loh, Lamperski, Full, and Cowan]{lee2008templates}
Jusuk Lee, Simon~N Sponberg, Owen~Y Loh, Andrew~G Lamperski, Robert~J Full, and Noah~J Cowan.
\newblock Templates and anchors for antenna-based wall following in cockroaches and robots.
\newblock \emph{IEEE Transactions on Robotics}, 24\penalty0 (1):\penalty0 130--143, 2008.

\bibitem[Lewinger et~al.(2005)Lewinger, Harley, Ritzmann, Branicky, and Quinn]{lewinger2005insect}
William~A Lewinger, Cynthia~M Harley, Roy~E Ritzmann, Michael~S Branicky, and Roger~D Quinn.
\newblock Insect-like antennal sensing for climbing and tunneling behavior in a biologically-inspired mobile robot.
\newblock In \emph{Proceedings of the 2005 IEEE International Conference on Robotics and Automation}, pages 4176--4181. IEEE, 2005.

\bibitem[Li et~al.(2024)Li, Xing, Khan, Zhong, and Cutkosky]{li2024whisker}
Hao Li, Chengyi Xing, Saad Khan, Miaoya Zhong, and Mark~R Cutkosky.
\newblock Whisker-inspired tactile sensing: A sim2real approach for precise underwater contact tracking.
\newblock \emph{arXiv preprint arXiv:2410.14005}, 2024.

\bibitem[Lobo et~al.(2019)Lobo, Nordbeck, Raja, Chemero, Riley, Jacobs, and Travieso]{lobo2019route}
Lorena Lobo, Patric~C Nordbeck, Vicente Raja, Anthony Chemero, Michael~A Riley, David~M Jacobs, and David Travieso.
\newblock Route selection and obstacle avoidance with a short-range haptic sensory substitution device.
\newblock \emph{International Journal of Human-Computer Studies}, 132:\penalty0 25--33, 2019.

\bibitem[Luneckas et~al.(2021)Luneckas, Luneckas, Udris, Plonis, Maskeli{\=u}nas, and Dama{\v{s}}evi{\v{c}}ius]{luneckas2021hybrid}
Mindaugas Luneckas, Tomas Luneckas, Dainius Udris, Darius Plonis, Rytis Maskeli{\=u}nas, and Robertas Dama{\v{s}}evi{\v{c}}ius.
\newblock A hybrid tactile sensor-based obstacle overcoming method for hexapod walking robots.
\newblock \emph{Intelligent Service Robotics}, 14:\penalty0 9--24, 2021.

\bibitem[Luo et~al.(2024)Luo, Li, Yu, Wang, Wu, and Zhu]{luo2024pie}
Shixin Luo, Songbo Li, Ruiqi Yu, Zhicheng Wang, Jun Wu, and Qiuguo Zhu.
\newblock Pie: Parkour with implicit-explicit learning framework for legged robots.
\newblock \emph{IEEE Robotics and Automation Letters}, 2024.

\bibitem[Miyake et~al.(2009)Miyake, Ishihara, and Tomino]{miyake2009vacuum}
Tohru Miyake, Hidenori Ishihara, and Tatsuya Tomino.
\newblock Vacuum-based wet adhesion system for wall climbing robots-lubricating action and seal action by the liquid.
\newblock In \emph{2008 IEEE International Conference on Robotics and Biomimetics}, pages 1824--1829. IEEE, 2009.

\bibitem[Miyamoto et~al.(2021)Miyamoto, Kinugasa, Amasaki, Osuka, Hayashi, and Yoshida]{miyamoto2021analysis}
Naoki Miyamoto, Tetsuya Kinugasa, Tatsuya Amasaki, Koichi Osuka, Ryota Hayashi, and Koji Yoshida.
\newblock Analysis of body undulation using dynamic model with frictional force for myriapod robot.
\newblock \emph{Artificial Life and Robotics}, 26:\penalty0 29--34, 2021.

\bibitem[Mongeau et~al.(2014)Mongeau, Demir, Dallmann, Jayaram, Cowan, and Full]{mongeau2014mechanical}
Jean-Michel Mongeau, Alican Demir, Chris~J Dallmann, Kaushik Jayaram, Noah~J Cowan, and Robert~J Full.
\newblock Mechanical processing via passive dynamic properties of the cockroach antenna can facilitate control during rapid running.
\newblock \emph{Journal of Experimental Biology}, 217\penalty0 (18):\penalty0 3333--3345, 2014.

\bibitem[Mongeau et~al.(2015)Mongeau, Sponberg, Miller, and Full]{mongeau2015sensory}
Jean-Michel Mongeau, Simon~N Sponberg, John~P Miller, and Robert~J Full.
\newblock Sensory processing within cockroach antenna enables rapid implementation of feedback control for high-speed running maneuvers.
\newblock \emph{The Journal of Experimental Biology}, 218\penalty0 (15):\penalty0 2344--2354, 2015.

\bibitem[Mrva and Faigl(2015)]{mrva2015tactile}
Jakub Mrva and Jan Faigl.
\newblock Tactile sensing with servo drives feedback only for blind hexapod walking robot.
\newblock In \emph{2015 10th International Workshop on Robot Motion and Control (RoMoCo)}, pages 240--245. IEEE, 2015.

\bibitem[Nagakubo and Hirose(1994)]{nagakubo1994walking}
Akihiko Nagakubo and Shigeo Hirose.
\newblock Walking and running of the quadruped wall-climbing robot.
\newblock In \emph{Proceedings of the 1994 IEEE International Conference on Robotics and Automation}, pages 1005--1012. IEEE, 1994.

\bibitem[Nayak and Pradhan(2014)]{nayak2014design}
Ankit Nayak and SK~Pradhan.
\newblock Design of a new in-pipe inspection robot.
\newblock \emph{Procedia Engineering}, 97:\penalty0 2081--2091, 2014.

\bibitem[Nilsson(1998)]{nilsson1998snake}
Martin Nilsson.
\newblock Snake robot-free climbing.
\newblock \emph{IEEE Control Systems Magazine}, 18\penalty0 (1):\penalty0 21--26, 1998.

\bibitem[Ozkan-Aydin et~al.(2020)Ozkan-Aydin, Chong, Aydin, and Goldman]{ozkan2020systematic}
Yasemin Ozkan-Aydin, Baxi Chong, Enes Aydin, and Daniel~I Goldman.
\newblock A systematic approach to creating terrain-capable hybrid soft/hard myriapod robots.
\newblock In \emph{2020 3rd IEEE International Conference on Soft Robotics (RoboSoft)}, pages 156--163. IEEE, 2020.

\bibitem[Parness et~al.(2017)Parness, Abcouwer, Fuller, Wiltsie, Nash, and Kennedy]{parness2017lemur}
Aaron Parness, Neil Abcouwer, Christine Fuller, Nicholas Wiltsie, Jeremy Nash, and Brett Kennedy.
\newblock Lemur 3: A limbed climbing robot for extreme terrain mobility in space.
\newblock In \emph{2017 IEEE international conference on robotics and automation (ICRA)}, pages 5467--5473. IEEE, 2017.

\bibitem[Pezzementi et~al.(2011)Pezzementi, Reyda, and Hager]{pezzementi2011object}
Zachary Pezzementi, Caitlin Reyda, and Gregory~D Hager.
\newblock Object mapping, recognition, and localization from tactile geometry.
\newblock In \emph{2011 IEEE International Conference on Robotics and Automation}, pages 5942--5948. IEEE, 2011.

\bibitem[Qi et~al.(2021)Qi, Lin, Hong, Chen, and Zhang]{qi2021perceptive}
Shuhao Qi, Wenchun Lin, Zejun Hong, Hua Chen, and Wei Zhang.
\newblock Perceptive autonomous stair climbing for quadrupedal robots.
\newblock In \emph{2021 IEEE/RSJ International Conference on Intelligent Robots and Systems (IROS)}, pages 2313--2320. IEEE, 2021.

\bibitem[Ramesh et~al.(2022)Ramesh, Fu, and Li]{ramesh2022sensnake}
Divya Ramesh, Qiyuan Fu, and Chen Li.
\newblock Sensnake: A snake robot with contact force sensing for studying locomotion in complex 3-d terrain.
\newblock In \emph{2022 International Conference on Robotics and Automation (ICRA)}, pages 2068--2075. IEEE, 2022.

\bibitem[Rashed et~al.(2019)Rashed, Ramzy, Vaquero, El~Sallab, Sistu, and Yogamani]{rashed2019fusemodnet}
Hazem Rashed, Mohamed Ramzy, Victor Vaquero, Ahmad El~Sallab, Ganesh Sistu, and Senthil Yogamani.
\newblock Fusemodnet: Real-time camera and lidar based moving object detection for robust low-light autonomous driving.
\newblock In \emph{Proceedings of the IEEE/CVF International Conference on Computer Vision Workshops}, pages 0--0, 2019.

\bibitem[Saunders et~al.(2006)Saunders, Goldman, Full, and Buehler]{saunders2006rise}
Aaron Saunders, Daniel~I Goldman, Robert~J Full, and Martin Buehler.
\newblock The rise climbing robot: body and leg design.
\newblock In \emph{Unmanned Systems Technology VIII}, volume 6230, pages 401--413. SPIE, 2006.

\bibitem[Song et~al.(2022)Song, Zhang, Meng, Chen, and Huang]{song2022gait}
Xingguo Song, Xiaolong Zhang, Xiangyin Meng, Chunjun Chen, and Dashan Huang.
\newblock Gait optimization of step climbing for a hexapod robot.
\newblock \emph{Journal of Field Robotics}, 39\penalty0 (1):\penalty0 55--68, 2022.

\bibitem[Spenko et~al.(2008)Spenko, Haynes, Saunders, Cutkosky, Rizzi, Full, and Koditschek]{spenko2008biologically}
Matthew~J Spenko, G~Clark Haynes, JA~Saunders, Mark~R Cutkosky, Alfred~A Rizzi, Robert~J Full, and Daniel~E Koditschek.
\newblock Biologically inspired climbing with a hexapedal robot.
\newblock \emph{Journal of field robotics}, 25\penalty0 (4-5):\penalty0 223--242, 2008.

\bibitem[Teder et~al.(2024)Teder, Chong, He, Wang, Iaschi, Soto, and Goldman]{teder2024effective}
Erik Teder, Baxi Chong, Juntao He, Tianyu Wang, Massimiliano Iaschi, Daniel Soto, and Daniel~I Goldman.
\newblock Effective self-righting strategies for elongate multi-legged robots.
\newblock \emph{arXiv preprint arXiv:2410.01056}, 2024.

\bibitem[Verma et~al.(2022)Verma, Kaiwart, Dubey, Naseer, and Pradhan]{verma2022review}
Ankush Verma, Ayush Kaiwart, Nikhil~Dhar Dubey, Farman Naseer, and Swastik Pradhan.
\newblock A review on various types of in-pipe inspection robot.
\newblock \emph{Materials Today: Proceedings}, 50:\penalty0 1425--1434, 2022.

\bibitem[Wu et~al.(2019)Wu, Huh, Sabin, Suresh, and Cutkosky]{wu2019tactile}
X~Alice Wu, Tae~Myung Huh, Aaron Sabin, Srinivasan~A Suresh, and Mark~R Cutkosky.
\newblock Tactile sensing and terrain-based gait control for small legged robots.
\newblock \emph{IEEE Transactions on Robotics}, 36\penalty0 (1):\penalty0 15--27, 2019.

\bibitem[Yan et~al.(1999)Yan, Shuliang, Dianguo, Yanzheng, Hao, and Xueshan]{yan1999development}
Wang Yan, Liu Shuliang, Xu~Dianguo, Zhao Yanzheng, Shao Hao, and Gao Xueshan.
\newblock Development and application of wall-climbing robots.
\newblock In \emph{Proceedings 1999 IEEE international conference on robotics and automation (Cat. No. 99CH36288C)}, volume~2, pages 1207--1212. IEEE, 1999.

\bibitem[Yasui et~al.(2017)Yasui, Sakai, Kano, Owaki, and Ishiguro]{yasui2017decentralized}
Kotaro Yasui, Kazuhiko Sakai, Takeshi Kano, Dai Owaki, and Akio Ishiguro.
\newblock Decentralized control scheme for myriapod robot inspired by adaptive and resilient centipede locomotion.
\newblock \emph{PloS one}, 12\penalty0 (2):\penalty0 e0171421, 2017.

\bibitem[Yim et~al.(2001)Yim, Homans, and Roufas]{yim2001climbing}
Mark Yim, Sam Homans, and Kimon Roufas.
\newblock Climbing with snake-like robots.
\newblock In \emph{IFAC workshop on mobile robot technology}, volume~1, page~2, 2001.

\bibitem[Ying et~al.(2017)Ying, Li, Ren, Wang, and Wang]{ying2017new}
Zhenqiang Ying, Ge~Li, Yurui Ren, Ronggang Wang, and Wenmin Wang.
\newblock A new low-light image enhancement algorithm using camera response model.
\newblock In \emph{Proceedings of the IEEE international conference on computer vision workshops}, pages 3015--3022, 2017.

\bibitem[Zamir et~al.(2021)Zamir, Arora, Khan, Khan, and Shao]{zamir2021learning}
Syed~Waqas Zamir, Aditya Arora, Salman Khan, Fahad~Shahbaz Khan, and Ling Shao.
\newblock Learning digital camera pipeline for extreme low-light imaging.
\newblock \emph{Neurocomputing}, 452:\penalty0 37--47, 2021.

\end{thebibliography}

\end{document}